\documentclass[10pt,twocolumn,letterpaper]{article}

\usepackage{cvpr}
\usepackage{times}
\usepackage{epsfig}
\usepackage{graphicx}
\usepackage{placeins}
\usepackage{amsmath}
\usepackage{amssymb}
\usepackage{paralist}
\usepackage{hhline}
\usepackage{multirow}
\usepackage{soul}
\usepackage{cancel}
\usepackage{booktabs}
\usepackage[ruled,vlined]{algorithm2e}

\SetCommentSty{mycommfont}
\usepackage[dvipsnames]{xcolor}

\newcommand{\cg}{\mathrm{CG}}

\newcommand{\dist}{\mathrm{dist}}
\newcommand{\gnn}{\mathrm{GNN}}
\newcommand{\cnn}{\mathrm{CNN}}

\newcommand{\troi}{\mathrm{RoI}}

\newcommand{\model}{\mathcal{M}}
\newcommand{\loss}{\mathcal{L}}
\newcommand{\entropy}{\mathcal{H}}

\newcommand{\kd}{\mathrm{KD}}
\newcommand{\ce}{\mathrm{CE}}
\newcommand{\deeplearning}{\mathrm{DL}}

\newcommand{\avg}{\text{avg}}
\newcommand{\corr}{\text{corr}}

\def\infinity{\rotatebox{90}{8}}

\DeclareMathOperator*{\argmax}{argmax} 
\SetKwInOut{Parameter}{Parameter}
\newcommand{\conceptset}{\mathcal{C}}
\newcommand{\attributeset}{\mathcal{A}}
\newcommand{\graphset}{\mathcal{G}}
\newcommand{\kset}{\mathcal{K}}
\newcommand{\tumorset}{\mathcal{T}}
\newcommand{\importanceset}{\mathcal{I}}
\newcommand{\dataset}{\mathcal{D}}

\newcommand{\gnnexplainer}{\textsc{GnnExplainer}}
\newcommand{\graphgradcam}{\textsc{GraphGrad-CAM}}
\newcommand{\graphgradcampp}{\textsc{GraphGrad-CAM++}}
\newcommand{\gradcam}{\textsc{Grad-CAM}}
\newcommand{\gradcampp}{\textsc{Grad-CAM++}}
\newcommand{\lrpfc}{\textsc{LRP-FC}}
\newcommand{\graphlrp}{\textsc{GraphLRP}}

\newcommand{\random}{\textsc{Random}}

\usepackage[pagebackref=true,breaklinks=true,letterpaper=true,colorlinks,bookmarks=false]{hyperref}

\cvprfinalcopy % *** Uncomment this line for the final submission

\def\cvprPaperID{8266} % *** Enter the CVPR Paper ID here

\ifcvprfinal\fi
\begin{document}

%%%%%%%%% TITLE
% \title{Explainable Graph Representations in Digital Pathology}
% \title{Taking the Mystique Out: Quantifying Explainers of Black Box Graph Neural Networks in Computational Pathology Without Bells and Whistles}
% \title{Taking the Mystique Out: Quantifying Explainers of \\Graph Neural Networks in Computational Pathology}
\title{Quantifying Explainers of Graph Neural Networks in Computational Pathology}

\author{
% authors:
% Option 1:
% Guillaume Jaume$^{*,1,2}$ \and
% Pushpak Pati$^{*,1,3}$ \and
% Behzad Bozorgtabar$^{1}$ \and
% Antonio Foncubierta$^{2}$ \and
% Anna Maria Anniciello$^{4}$ \and
% Florinda Feroce$^{4}$ \and
% Tilman Rau$^{5}$ \and
% Jean-Philippe Thiran$^{1}$ \and
% Maria Gabrani$^{2}$ \and
% Orcun Goksel$^{3,6}$\\
% Option 2:
Guillaume Jaume$^{1,2}$\thanks{denotes equal contribution},
Pushpak Pati$^{1,3,*}$,
Behzad Bozorgtabar$^{2}$,\\
Antonio Foncubierta$^{1}$,
Anna Maria Anniciello$^{4}$,
Florinda Feroce$^{4}$,
Tilman Rau$^{5}$,\\
Jean-Philippe Thiran$^{2}$,
Maria Gabrani$^{1}$,
Orcun Goksel$^{3,6}$\\
% affiliations:
% Option 1:
% $^1$Signal Processing Laboratory 5, EPFL, Lausanne, Switzerland\\
% $^2$IBM Zurich Research Lab, Zurich, Switzerland\\
% $^3$Computer-Assisted Applications in Medicine, ETH Zurich, Zurich, Switzerland\\
% $^4$National Cancer Institute - IRCCS-Fondazione Pascale, Naples, Italy\\
% $^5$Institute of Pathology of the University Bern, Bern, Switzerland\\
% $^6$Department of Information Technology, Uppsala University, Sweden\\
% Option 2:
$^1$IBM Research Zurich,
$^2$EPFL Lausanne,
$^3$ETH Zurich,\\
$^4$Fondazione Pascale,
$^5$University of Bern,
$^6$Uppsala University\\
% emails (@TODO: not include, only 2m all of them?)
{\tt\small \{gja,pus\}@zurich.ibm.com} 
}
\maketitle

%------------------------------------------------------------------------------------------
% ABSTRACT
%------------------------------------------------------------------------------------------

\begin{abstract}
Explainability of deep learning methods is imperative to facilitate their clinical adoption in digital pathology. However, popular deep learning methods and explainability techniques (explainers) based on pixel-wise processing disregard biological entities' notion, thus complicating comprehension by pathologists. 
In this work, we address this by adopting biological entity-based graph processing and graph explainers enabling explanations accessible to pathologists. In this context, a major challenge becomes to discern meaningful explainers, particularly in a standardized and quantifiable fashion. To this end, we propose herein a set of novel quantitative metrics based on statistics of class separability using pathologically measurable concepts to characterize graph explainers. We employ the proposed metrics to evaluate three types of graph explainers, namely the layer-wise relevance propagation, gradient-based saliency, and graph pruning approaches, to explain Cell-Graph representations for Breast Cancer Subtyping. The proposed metrics are also applicable in other domains by using domain-specific intuitive concepts. We validate the qualitative and quantitative findings on the BRACS dataset, a large cohort of breast cancer $\troi$s, by expert pathologists. The code, data, and models can be accessed here\footnote{\url{https://github.com/histocartography/patho-quant-explainer}}.

\end{abstract}

%------------------------------------------------------------------------------------------
% INTRODUCTION
%------------------------------------------------------------------------------------------
\vspace{-0.2em}
\section{Introduction}
\label{sec:introduction}
% Introduce deep neural network in digital pathology. Need for explainability in digital pathology
Histopathological image understanding has been revolutionized by recent machine learning advancements, especially deep learning ($\deeplearning$) \cite{Bera2019,Serag2019}. $\deeplearning$ has catered to increasing diagnostic throughput as well as a need for high predictive performance, reproducibility and objectivity.
% quality quantitative assessments in histopathological diagnoses.
However, such advantages come at the cost of a reduced transparency in decision-making processes~\cite{Holzinger2017,Tizhoosh2018,Hagele2020}. Considering the need for reasoning any clinical decision, it is imperative to enable the explainability of $\deeplearning$ decisions to pathologists.
%for adoption in clinical practice.

\begin{figure}[!t]
\centering
\centerline{\includegraphics[width=\linewidth]{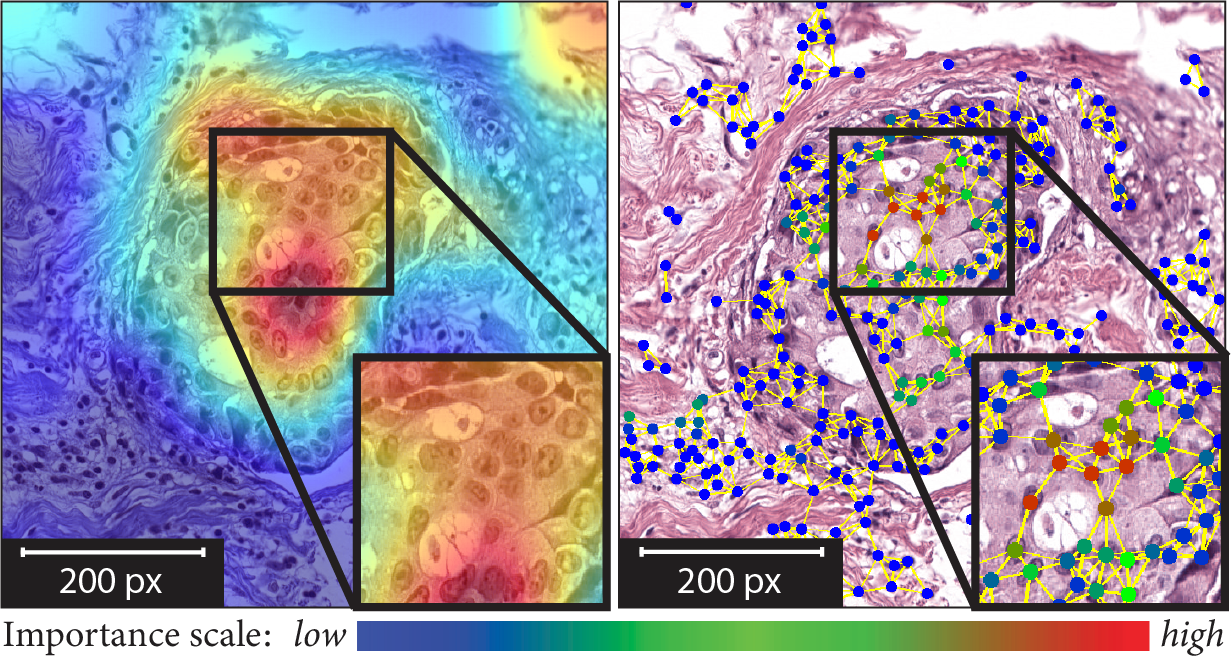}}
\caption{Sample explanations produced by pixel- and entity-based explainability techniques for a ductal carcinoma \emph{in situ} $\troi$.}
\vspace{-1em}
\label{fig:teaser}
\end{figure}

% Introduce interpretability in CNN, visual explanation. Limitations of CNN and visual explanation.
Inspired by the explainability techniques (explainers) for $\deeplearning$ model decisions on natural images \cite{Simonyan2013,Zeiler2014,Yosinski2015,Bach2015,Montavon2015,Selvaraju2017,Kindermans2017,Zintgraf2017, Chattopadhay2018,Kim2018}, several explainers have been implemented in digital pathology, such as feature attribution \cite{Korbar2017, Binder2018, Hagele2020}, concept attribution \cite{Graziani2020}, and attention-based learning \cite{Lu2020b}.
% FRA: Add a citation to our ICML-w paper. It should not be an obvious self-reference, neither an obvious omission.
However, pixel-level explanations, exemplified in  Figure~\ref{fig:teaser}, pose several notable issues, including:
(1)~a pixel-wise analysis disregards the notion of biological tissue entities, their topological distribution, and inter-entity interactions;
(2)~a typical patch-based $\deeplearning$ processing and explainer fail to accommodate complete tumor macro-environment information; and
(3)~pixel-wise visual explanations %, \ie heatmaps of salient regions, 
tend to be blurry.
% Introduce Entity Graphs, Advantages of GNN over CNN, Advantage of GNN explanation over visual explanation
Explainability in entity space is thus a natural choice to address the above issues. To that end, an entity graph representation is built for a histology image, where nodes and edges denote biological entities and inter-entity interactions followed by a Graph Neural Network ($\gnn$) \cite{Kipf2017,Xu2018b}. The choice of entities, such as cells \cite{Gunduz2004,Zhou2019,Pati2020}, tissues \cite{Pati2020} or others, can be task-dependent. Subsequently, explainers for graph-structured data \cite{Baldassarre2019,Pope2019,Ying2019} applied to the entity graphs highlight responsible entities for the concluded diagnosis, thereby generating intuitive explanations for pathologists.

% Metrics
%In the presence of various graph explainers producing distinct explanations, it is crucial to assess explainer quality \cite{Arrieta2020}. Qualitative assessment requires retrospective requests to expert pathologists; thus, imparting analytical subjectivity and financial overhead. Further, the explanations are difficult to be comprehensively analyzed by pathologists, as they are not expressed using pathological terms, and are incongruous with prior pathological knowledge. Therefore, pathologically congruent quantitative assessment of explanations is essential. Existing evaluation measures pertaining to quantitative assessment evaluate explainer \emph{fidelity} \cite{Ribeiro2016,Dhurandhar2017,Samek2017,Hoffman2018,Mohseni2018,Pope2019}, $\ie$ measuring model correctness for inputting explanation. \textit{Fidelity} is implicitly preserved in graph explainers, and is not comprehensive from a pathological standpoint. Therefore, we propose a set of novel quantitative metrics based on statistics of class separability using pathologically understandable concepts. 
%The proposed metrics can accommodate various types of graph explainers and different types of concepts ($\ie$ discrete and continuous). However, the metrics are applicable to other domains by incorporating domain-specific concepts.

In the presence of various graph explainers producing distinct explanations for an input, it is crucial to discern the explainer that best fits the explainability definition \cite{Arrieta2020}. In the context of computational pathology, explainability is defined as making the $\deeplearning$ decisions understandable to pathologists \cite{Holzinger2017}. To this end, the qualitative evaluation of explainers' explanations by pathologists is the candid measure. However, it requires evaluations by task-specific expert pathologists, which is subjective, time-consuming, cumbersome, and expensive.
% But, such ground truth is unavailable or infeasible in other domains, \eg medical imaging.
% @TODO: we say that they are difficult to be analysed by pathologists and before we were saying that entity graph explanations are intuitive. This is a contradiction. 
Additionally, though the explanations are intuitive, they do not relate to pathologist-understandable terminologies, \eg ``How big are the important nuclei?", ``How irregular are their shape?" etc., which toughens the comprehensive analysis.
These bottlenecks undermine not only any qualitative assessment but also quantitative metrics requiring user interactions~\cite{Mohseni2018}.
Furthermore, expressing the quantitative metrics in user-understandable terminologies \cite{Arrieta2020} is fundamental to achieve interpretability \cite{Doshivelez2017, Nguyen2020}.
%Indeed it has been argued that the choice of relevant \emph{units of explanations} is fundamental to achieve interpretability \cite{Doshivelez2017, Nguyen2020}.
To this end, the most popular quantitative metric, explainer \emph{fidelity}~\cite{Ribeiro2016,Dhurandhar2017,Samek2017,Hoffman2018,Mohseni2018,Pope2019}, is not satisfactory. Moreover, explainers intrinsically maintain high-\emph{fidelity}, \eg $\gnnexplainer$~\cite{Ying2019} produces an explanation to match the $\gnn$'s prediction on the original graph.

Thus, we propose a set of novel user-independent quantitative metrics expressing pathologically-understandable \emph{concepts}. The proposed metrics are based on class separability statistics using such \emph{concepts}, and they are applicable in other domains by incorporating domain-specific \emph{concepts}.
% Contribution
We use the proposed metrics to evaluate three types of graph-explainers, (1) graph pruning: $\gnnexplainer$ \cite{Ying2019, Jaume2020}, (2) gradient-based saliency: $\graphgradcam$ \cite{Selvaraju2017,Pope2019}, $\graphgradcampp$ \cite{Chattopadhay2018}, (3) layer-wise relevance propagation: $\graphlrp$~\cite{Bach2015,Montavon2015,Schwarzenberg2019}, for explaining Cell-Graphs \cite{Gunduz2004} in Breast Cancer Subtyping as shown in Figure \ref{fig:teaser}.
Our specific contributions in this work are:
\begin{itemize}
    \item A set of novel quantitative metrics based on the statistics of class separability using domain-specific \emph{concepts} to characterize graph explainability techniques. To the best of our knowledge, our metrics are the first of their kind to quantify explainability based on domain-understandable terminologies;
    \item Explainability in computational pathology using pathologically intuitive entity graphs; 
    \item Extensive qualitative and quantitative assessment of various graph explainability techniques in computational pathology, with a validation of the findings by expert pathologists.
\end{itemize}

\begin{figure*}[!t]
\centering
\centerline{\includegraphics[width=\textwidth]{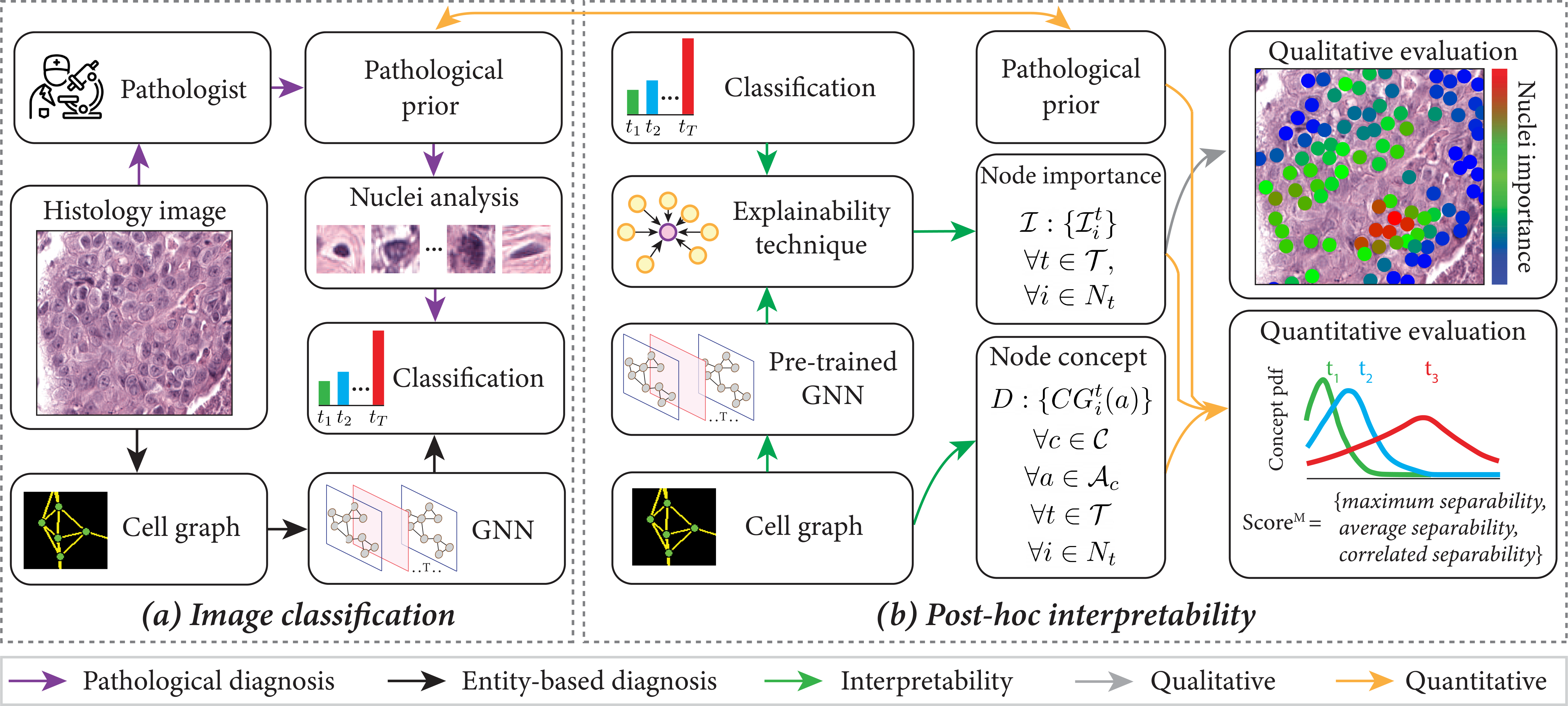}}
\caption{Overview of the proposed framework. 
(a) presents pathologist, and entity-based (cell-graph + $\gnn$) diagnosis of a histology image. 
(b) presents nuclei-level pathologically relevant \emph{concept} measure $D$, a post-hoc graph explainability technique to derive nuclei-level importance $\importanceset$ for \emph{concepts} $\conceptset$, measurable \emph{attributes} $\attributeset_c$, and classes $\tumorset$. $D$, $\importanceset$ and prior pathological knowledge defining \emph{concepts'} relevance are utilized to propose a novel set of quantitative metrics to evaluate the explainer quality in  pathologist-understandable terms.}
\label{fig:approach} 
\end{figure*}

%------------------------------------------------------------------------------------------
% RELATED WORK
%------------------------------------------------------------------------------------------
\section{Related work}

\textbf{Graphs in Digital Pathology:}
Graph-based tissue image analysis effectively describes a tissue environment by incorporating morphology, topology, and tissue components interactions.
To this end, cell-graph ($\cg$) is the most popular graph representation, where nodes and edges depict cells and cellular interactions \cite{Gunduz2004}. Cell morphology is embedded in the nodes via hand-crafted features~\cite{Gunduz2004,Zhou2019,Pati2020} or $\deeplearning$ features \cite{Chen2019,Pati2021}. The graph topology is heuristically defined using k-Nearest Neighbors, probabilistic modeling, Waxman model etc. \cite{Sharma2015} 
Subsequently, the $\cg$s are processed by classical machine learning \cite{Sharma2015,Sharma2016,Sharma2017} or $\gnn$ \cite{Zhou2019,Chen2019,Gadiya2020,Pati2020} to map the tissue structure to function relationship.
Recently, improved graph-representations using patches \cite{Aygunes2020}, tissue components \cite{Pati2020}, and hierarchical cell-to-tissue relations \cite{Pati2020} are proposed to enhance the  structure-function mapping.
Other graph-based applications in computational pathology include cellular community detection \cite{Javed2020}, whole-slide image classification \cite{Zhao2020, Adnan2020} etc. Intuitively, a graph representation utilizes pathologically relevant entities to represent a tissue specimen, which allows pathologists to readily relate with the input, also enabling them to include any task-specific prior knowledge.

\textbf{Explainability in Digital Pathology:}
Explainability is an integral part of pathological diagnosis. Though $\deeplearning$ solutions have achieved remarkable diagnostic performance, their lack of explainability is unacceptable in the medical community~\cite{Tizhoosh2018}. Recent studies have proposed visual explanations~\cite{Hagele2020} and salient regions~\cite{Korbar2017,Hagele2020} using feature-attribution techniques \cite{Selvaraju2017,Chattopadhay2018}.
% \gja{However, the resulting pixel-level explanations are often blurry and do not emphasize on the biological entities.}
%such pixelated and blurry explanations have not been comprehensible to pathologists as they ignore biological entities' notion. 
Differently, concept-attribution technique~\cite{Graziani2020} evaluates the sensitivity of network output w.r.t.\ quantifiable image-level pathological \emph{concepts} in patches. Although such explanations are pathologist-friendly, image-level \emph{concepts} are neither fit nor meaningful for real-world large histology images that contain many localized concepts.
Furthermore, attention-based learning~\cite{Lu2020b}, and multimodal mapping between image and diagnostic report~\cite{Zhang2019} are devised to localize network attention. 
However, the pixel-wise and patch-based processing in all the aforementioned techniques ignore biological entities' notion; thus, they are not easily understood by pathologists.
Separately, the earlier stated entity graph-based processing provides an intuitive platform for pathologists. However, research on explainability and visualization using entity graphs has been scarce: CGC-Net~\cite{Zhou2019} analyzes cluster assignment of nodes in $\cg$ to group them according to their appearance and tissue types. CGExplainer~\cite{Jaume2020} introduces a post-hoc graph-pruning explainer to identify decisive cells and interactions. Robust spatial filtering~\cite{Sureka2020} utilizes an attention-based $\gnn$ and node occlusion to highlight cell contributions.
No previous work has comprehensively analyzed and quantified graph explainers in computational pathology while expressing explanations in a pathologist-understandable form to the best of our knowledge. 
This gap between the existing and desired explainability of $\deeplearning$ outputs in digital pathology motivates our work herein.
%Thus, extensive comparison and quantification of explainability techniques is crucial in computational pathology.

%------------------------------------------------------------------------------------------
% METHOD
%------------------------------------------------------------------------------------------
\section{Method}
% Introduction of the methodology
In this section, we present entity graph processing, explainability methods, and our proposed evaluation metrics. First, we transform a histology region-of-interest ($\troi$) into a \emph{biological entity graph}. Second, we introduce a ``black-box" $\gnn$ that maps the \emph{entity graph} to a corresponding class label. Third, we employ a post-hoc graph explainer to generate explanations. Finally, we perform a qualitative and quantitative assessments of the generated explanations. An overview of the methodology is shown in Figure~\ref{fig:approach}.

\subsection{Entity graph notations}
We define an attributed undirected entity graph $G := (V, E, H)$ as a set of nodes $V$, edges $E$, and node attributes $H \in \mathbb{R}^{|V| \times d}$. $d$ denotes the number of attributes per node, and $|.|$ denotes set cardinality.
%The entity-graph is undirected, $\ie$ an edge between two nodes $u, v \in V$ is \mbox{$e_{uv} \in E \iff e_{vu} \in E$}.
The graph topology is defined by a symmetric graph adjacency, $A \in \mathbb{R}^{|V| \times |V|}$, where $A_{u,v} = 1$ if $e_{uv} \in E$.
We denote the neighborhood of a node $v \in V$ as $\mathcal{N}(v) := \{u \in V \; | \; v \in V, \; e_{uv} \in E \;\}$. 
We denote a set of graphs as $\graphset$.
%$:= \{G_0, G_1, ..., G_{|\graphset|-1}\}$.

\subsection{Entity graph construction} \label{sec:graph_construction}
% pixel to entity shift
Our methodology begins with transforming $\troi$s into entity graphs. It ensures the method's inputs are pathologically interpretable, as the inputs consist of biologically-defined objects that pathologists can directly \emph{relate-to} and \emph{reason-with}. Thus, image-to-graph conversion moves from $\textit{uninterpretable}$ to $\textit{interpretable}$ input space.
In this work, we consider cells as entities, thereby transforming $\troi$s into cell-graphs ($\cg$s). A $\cg$ nodes and edges capture the morphology of cells and cellular interactions. A $\cg$ topology acquires both tissue micro and macro-environment, which is crucial for characterizing cancer subtypes.

% node and node features 
First, we detect nuclei in a $\troi$ at 40$\times$ magnification using Hover-Net~\cite{Graham2019}, a nuclei segmentation algorithm pre-trained on MoNuSeg~\cite{Kumar2017}. We process patches of size 72$\times$72 around the nuclei by ResNet34 \cite{He2016} pre-trained on ImageNet~\cite{JiaDeng2009} to produce nuclei visual attributes. We further concatenate nuclei spatial attributes, $\ie$  nuclei centroids min-max normalized by $\troi$ dimension. 
The nuclei and their attributes (visual and spatial) define the nodes and node attributes of the $\cg$, respectively.
% cell-graph construction: edges 
Following prior work~\cite{Pati2020}, we construct the $\cg$~topology by employing thresholded $\textit{k}$-Nearest Neighbors algorithm. We set $\textit{k}=5$, and prune the edges longer than $50$ pixels (12.5 $\mu$m).
% given the scanner resolution of 0.25 $\mu$m/pixel).
%FRA: I edited this:
% The $\cg$~topology follows the heuristic that, two nearby nuclei in a tissue are more likely to interact, therefore, they should form an edge in the $\cg$.
% To this
The $\cg$-topology encodes how likely two nearby nuclei will interact~\cite{Francis1997}. 
% follows the idea that, two nearby nuclei in a tissue are more likely to interact~\cite{Francis1997}, therefore, they should form an edge in the $\cg$.
A $\cg$ example is presented in Figure~\ref{fig:teaser}.
%The $\cg$~topology is defined heuristically, $\ie$ two nearby nuclei in a tissue are more likely to interact, thus, they should form an edge in the $\cg$. To this end, we employ $\textit{k}$-Nearest Neighbors ($\textit{k}$-NN) algorithm, \ie given a node $v$, an edge $e_{uv}$ is formed if $u$ belongs to the $k$ nearest nodes of $v$. To accommodate the isolated nuclei in the tissue, we threshold the $\textit{k}$-NN graph by a minimum distance. Following prior work~\cite{Pati2020}, we set $\textit{k}=5$, and prune the edges lengthier than $50$ pixels (12.5 $\mu$m given the scanner resolution of 0.25 $\mu$m/pixel). A $\cg$ example is presented in Figure~\ref{fig:teaser}.

\subsection{Entity graph learning} \label{sec:graph_learning}
Given $\graphset$, the set of $\cg$s, the aim is to infer the corresponding cancer subtypes. We use $\gnn$s~\cite{Scarselli2009, Defferrard2016, Kipf2017, Hamilton2017b, Velickovic2017, Ying2018, Gilmer2017}, the conceptual analogous of 2D convolution for graph-structured data, to classify the $\cg$s.
% $\gnn$s can be formulated using the Message Passing Neural Network (MPNN) framework~\cite{Gilmer2017}.
% A GNN iteratively updates the state of each node using the neighboring node states to contextualise the node representation.
% Intuitively, the state of each node is updated iteratively using the neighboring node states to contextualise the node representation.
% Formally, 
A $\gnn$ layer follows two steps: for each node $v \in V$, \begin{inparaenum}[(i)]
\item \emph{aggregation step}: the states of neighboring nodes, $\mathcal{N}(v)$, are aggregated via a differentiable and permutation-invariant operator to produce $a(v) \in \mathbb{R}^d$, then,
\item \emph{update step}: the state of $v$ is updated by combining the current node state $h(v) \in \mathbb{R}^d$ and the aggregated message $a(v)$ via another differentiable operator. \end{inparaenum} 
After $L$ iterations, \ie the number of $\gnn$ layers, a \emph{readout step} is employed to merge all the node states via a differentiable and permutation-invariant function to result in a fixed-size graph embedding.
Finally, the graph embeddings are processed by a classifier to predict the class label.

In this work, we use a flavor of Graph Isomorphism Network (GIN)~\cite{Xu2018b},
% a popular instance of MPNN,
that uses \emph{mean} and a \emph{multi-layer perceptron} (MLP) in the \emph{aggregation} and \emph{update} step respectively. Formally, we define a layer as,
\begin{align}
    h(v)^{(l+1)} = \mbox{MLP}^{(l)} \Big(h(v)^{(l)} + \frac{1}{|\mathcal{N}(v)|} \sum_{u \in \mathcal{N}(v)} h(u)^{(l)} \Big)
\end{align}
where $h(v)$ denotes features of node $v$, and $l \in \{1, ..., L\}$. Our $\gnn$ consists of 3-GIN layers, with each layer including a 2-layer MLP. The dimension of latent node embeddings is fixed to 64 for all layers. We use \emph{mean} operation in \emph{readout step}, and feed the graph embedding to a 2-layer MLP classifier. The $\gnn$ is trained end-to-end by minimizing cross-entropy loss between predicted logits and target cancer subtypes.
% linking this paragraph to Fig 1.
We emphasize that the entity-based processing follows a pathologist's diagnostic procedure that identifies diagnostically relevant nuclei and analyzes cellular morphology and topology in a $\troi$, as shown in Figure~\ref{fig:approach}. 

\begin{figure*}[!t]
\centering
\centerline{\includegraphics[width=\linewidth]{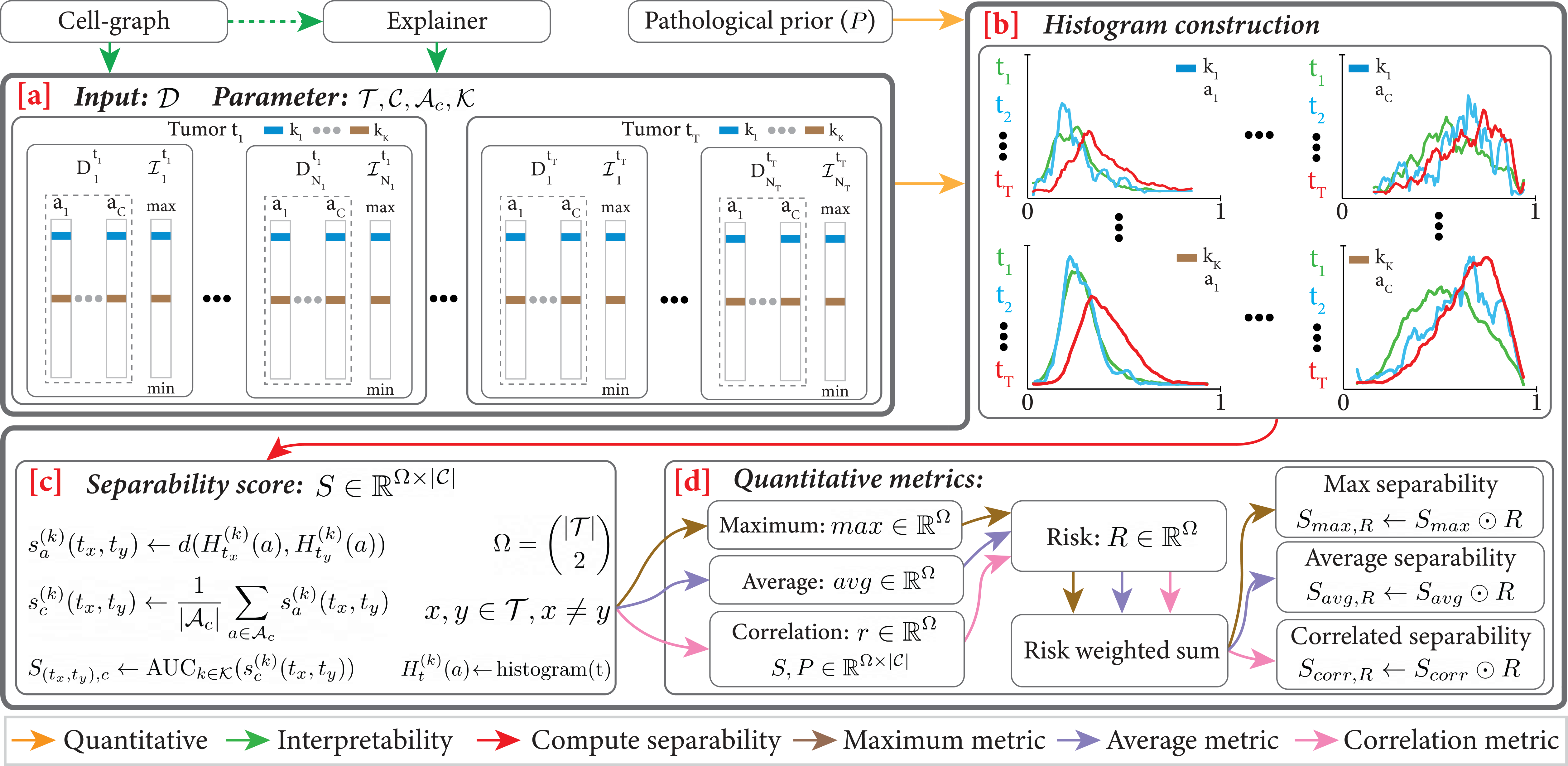}}
\caption{Overview of proposed quantitative assessment. (a) presents input dataset $\dataset$, and parameters \emph{concepts} $\conceptset$, measurable \emph{attributes} $\attributeset_c$, classes $\tumorset$, and importance thresholds $\kset$. For simplicity $|\attributeset_c|=1, \forall c \in \conceptset$ in this figure. (b) shows histogram probability densities for $\forall a \in \attributeset_c, \forall k \in \kset, \forall t \in \tumorset$. (c) displays the algorithm for computing class separability score $S$. (d) presents the algorithm for computing the proposed class separability-based risk-weighted quantitative metrics.}
\label{fig:metric}
\end{figure*}

\subsection{Post-hoc graph explainer} \label{sec:explainer}
% why do we want an explanation ?
We generate an explanation per entity graph by employing post-hoc graph explainers. The explanations allow to evaluate the pathological relevance of black-box neural network reasoning. Specifically, we aim to evaluate the agreement between the pathologically relevant set of nuclei in a $\troi$, and the explainer identified set of important nuclei, \ie nuclei driving the prediction, in corresponding $\cg$. In this work, we consider three types of graph explainers for explaining $\cg$s, which follow similar operational setting, \ie \begin{inparaenum}[(i)]
\item input data are attributed graphs, 
\item a $\gnn$ is trained \emph{a priori} to classify the input data, and
\item each data point can be inferred independently to produce an explanation.
\end{inparaenum}
We present the graph explainers in the following sections and their detailed mathematical formulations in the Appendix.

% explainability methods:
\textbf{$\graphlrp$:} Layerwise relevance propagation (LRP)~\cite{Bach2015} propagates the output logits backward in the network using a set of propagation rules to quantify the positive contribution of input pixels for a certain prediction. Specifically, LRP assigns an importance score to each neuron such that the output logit relevance is preserved across layers.
While initially developed for explaining fully-connected layers, LRP can be extended to $\gnn$ by treating the $\gnn$ \emph{aggregation step} as a fully connected layer that projects the graph adjacency matrix on the node attributes as in \cite{Schwarzenberg2019}. LRP outputs per-node importance.

\textbf{$\graphgradcam$:} $\gradcam$ \cite{Selvaraju2017} is a feature attribution approach designed for explaining $\cnn$s operating on images. It produces class activation explanation following two steps. First, it assigns weights to each channel of a convolutional layer $l$ by computing the gradient of the targeted output logit w.r.to each channel in layer $l$. Second, importance of the input elements are computed by the weighted combination of the forward activations at each channel in layer $l$. 
The extension to $\gnn$ is straightforward~\cite{Pope2019}, and only requires to compute the gradient of the predicted logits w.r.to a $\gnn$ layer. Following prior work~\cite{Pope2019}, we take the average of node-level importance-maps obtained from all the $\gnn$ layers $l \in \{1, ..., L\}$ to produce smooth per-node importance.

\textbf{$\graphgradcampp$:} $\gradcampp$ \cite{Chattopadhay2018} is an increment on $\gradcam$ by including spatial contributions into the channel-wise weight computation of a convolutional layer. The extension allows weighting the contribution by each spatial location at a layer for improved spatial localization. The spatial locations in a convolutional layer are analogous to the size of the graph in a $\gnn$ layer. With this additional consideration, we propose an extension of $\gradcampp$ to graph-structured data.

\textbf{$\gnnexplainer$:} $\gnnexplainer$~\cite{Ying2019, Jaume2020} is a graph pruning approach that aims to find a compact sub-graph $G_s \subset G$ such that mutual information between $G_s$ and $\gnn$ prediction of $G$ is maximized.  
Sub-graph $G_s$ is regarded as the explanation for the input graph $G$. $\gnnexplainer$ can be seen as a feature attribution technique with binarized node importances. To address the combinatorial nature of finding $G_s$, $\gnnexplainer$ formulates it as an optimization problem that learns a mask to activate or deactivate parts of the graph. \cite{Jaume2020} reformulates the initial approach in \cite{Ying2019} to learn a mask over the nodes instead of edges. The approach in~\cite{Jaume2020} is better suited for pathology as the nodes, \ie biological entities, are more intuitive and substantial for disease diagnosis than heuristically-defined edges.
% \begin{inparaenum}[(i)]
% \item minimize\behzad{s} the size of the mask,
% \item produce\behzad{s} a binary mask by minimizing element-wise masked entropy, and
% \item retain\behzad{s} the class prediction for using $G$. 
% \end{inparaenum} 
%The loss is optimized for each entity graph to result in per-node importance. 
The optimization for an entity graph results in per-node importance.

\subsection{Quantitative metrics for graph explainability} \label{sec:metrics}
In the presence of several graph explainers producing distinct explanations for an input, it is imperative to discern the explainer that produces the most pathologically-aligned explanation. Considering the limitations of existing qualitative and quantitative measures presented in Section \ref{sec:introduction}, we propose a novel set of quantitative metrics based on class separability statistics using pathologically relevant \emph{concepts}. Intuitively, a good explainer should emphasize the relevant \emph{concepts} that maximize the class separation. Details of the metric evaluations are presented as follows.

% Begin the quantitative metric
% INPUT:
\textbf{Input:} A graph explainer outputs an explanation, \ie node-level \emph{importance} $\mathcal{I}$, for an input $\cg$. To quantify a \emph{concept} $c \in \mathcal{C}$, $\conceptset$ denoting the set of \emph{concepts}, we measure nuclear \emph{attributes} $a \in \attributeset_{c}$ for each nucleus in $\cg$, \eg, for $c\!=\!$ \emph{nuclear shape}, we measure $\attributeset_{c}\!=\!$ \{\emph{perimeter, roughness, eccentricity, circularity}\}.
% re-formulate the introduction of D
% before:
% Given cancer subtypes $\tumorset$, and $N_t$ number of $\cg$s for $t\in \tumorset$, we obtain dataset $\dataset=\{(D_i^t, \importanceset_i^t) \}$, where $(D_i^t, \importanceset_i^t)$ denote the node-level \emph{attribute} matrix and \emph{importance} matrix for a $\cg$. 
% now:
We create a dataset $\dataset = \bigcup_{t \in \tumorset} \mathcal{D}_t$, $\tumorset$ denoting the set of cancer subtypes. We define $\mathcal{D}_t := \{(D^t_i, \mathcal{I}^t_i) | i = 1, \dots, N_t\} \forall t \in \tumorset$,  where $N_t$ is the number of CGs for tumor type $t$. $\mathcal{I}^t_i$ and $D^t_i$ are, respectively, the sorted importance matrix for a CG indexed by $i$ and corresponding node-level attribute matrix.
To perform inter-concept comparisons, we conduct \emph{attribute}-wise normalization across all $D^t_i \; \forall t,i$. In order to compare different explainers, we conduct $\cg$-wise normalization of $\importanceset$. The structure of input dataset $\dataset$ is presented in Figure \ref{fig:approach}\textcolor{red}{(a)}.

% PARAMETERS:
% Notably, the distinct underlying mechanism of different explainers assign node-level importances differently for a $\cg$.
% Further, the number of nodes vary across $\cg$s.
Note that the notion of important nuclei vary
\begin{inparaenum}[(1)]
\item per-$\cg$ since the number of nodes vary across $\cg$s, and
\item per-explainer.
\end{inparaenum}
%Hence, a \emph{fixed} threshold for selecting important nuclei across $\cg$s and explainers is not meaningful. To overcome this issue, we assess nuclei across $\cg$s and explainers by applying a set of thresholds $k \in \kset$ on the node-importances.
% Finally, we aggregate the assessments of an explainer for $k \in \kset$.
Hence, selecting a \emph{fixed} number of important nuclei per-$\cg$ and per-explainer is not meaningful. To overcome this issue, we assess different number of important nuclei $k \in \kset$, selected based on node importances, per-$\cg$ and per-explainer.
In the following sections we will show how to aggregate the results for a given explainer.

% HISTOGRAM:
\textbf{Histogram construction:} 
Given the input dataset $\dataset$, and parameters $\kset, \conceptset, \attributeset_c, \tumorset$, we apply threshold $k \in \kset$ on $\importanceset_i^t, \forall t \in \tumorset, \forall i \in N_t$ to select $\cg$-wise most important nuclei. The cancer subtype-wise selected set of nuclei data from $\dataset$ are used to construct histograms $H_t^{(k)}(a), \forall a \in \attributeset_c$, $\forall c \in \conceptset$ and $\forall t \in \tumorset$. For  histogram $H_t^{(k)}(a)$, bin-edges are decided by quantizing the complete range of \emph{attribute} $a$, \ie $\dataset(a)$, by a fixed step size. We convert each $H_t^{(k)}(a)$ into a probability density function.
Similarly, sets of histograms are constructed by applying different thresholds $k \in \kset$. Sample histograms are shown in Figure \ref{fig:approach}\textcolor{red}{(b)}.

\textbf{Separability Score (\emph{S}):} 
% We use $H_t^{(k)}(a), \forall t \in \tumorset$ to evaluate the pair-wise class separability.
Given two classes $t_x, t_y \in \tumorset$ and corresponding probability density functions $H_{t_x}^{(k)}(a)$ and $H_{t_y}^{(k)}(a)$, we compute \emph{class separability} $s_a^{(k)}(t_x, t_y)$ based on optimal transport as the Wasserstein distance between the two density functions. 
% replace this sentence:
% We use $s_a^{(k)}(t_x, t_y), \forall a \in \attributeset_c$, and aggregate them to obtain $s_c^{(k)}(t_x, t_y)$.
% by:
We average $s_a^{(k)}(t_x, t_y)$ over all $a \in \attributeset_c$ to obtain a score $s_c^{(k)}(t_x, t_y)$ for \emph{concept} $c$ and threshold $k$.
% replace:
% Further, we obtain $s_c^{(k)}(t_x, t_y) \; \forall \; k \in \kset$, and compute the area-under-the-curve (AUC) to get the aggregated class separability $S_{(t_x, t_y), c}$ for \emph{concept} $c$.
% by:
Finally, we compute the area-under-the-curve (AUC) over the threshold range $\mathcal{K}$ to get the aggregated class separability $S_{(t_x, t_y), c}$ for a \emph{concept} $c$.
The class separability score indicates the significance of \emph{concept} $c$ for the purpose of separating $t_x$ and $t_y$.
Thus, separability scores can be used to compare different \emph{concepts} and to identify relevant ones for differentiating $t_x$ and $t_y$. 
A pseudo-algorithm is presented in Algorithm \ref{algorithm:metric}, and illustrated in Figure \ref{fig:approach}\textcolor{red}{(c)}.
A separability matrix $S \in \mathbb{R}^{\Omega \times |\conceptset|}$ is built by computing class separability scores for all pair-wise classes, \ie $\forall \; (t_x, t_y) \in \Omega:= {|\tumorset| \choose 2}$ and $\forall c \in \conceptset$.

\textbf{Statistics of Separability Score:}
Since explainability is not uniquely defined, we include multiple metrics highlighting different facets.
We compute three separability statistics $\forall (t_x, t_y) \in \Omega$ using $S$ as given in Equation~\eqref{eqn:stats}, \ie
\begin{inparaenum}[(1)]
\item \emph{maximum}: the utmost separability,
\item \emph{average}: the expected separability. These two metrics encode (model+explainer)'s focus, \ie ``how much the black-box model implicitly uses the \emph{concepts} for class separability?" 
\item \emph{correlation}: encodes the agreement between (model+explainer)'s focus and pathological prior $P$. $P \in \mathbb{R}^{\Omega \times |\conceptset|}$ signifies the relevance $\forall c \in \conceptset$ for differentiating $(t_x, t_y) \in \Omega$, \eg nuclear \emph{size} is highly relevant for classifying benign and malignant tumor as important nuclei in malignant are larger than important nuclei in benign. 
\end{inparaenum}
\begin{align}
\label{eqn:stats}
\centering
\begin{aligned}
    s_{\max}(t_x,t_y) &= \max_{c \in \conceptset}S_{(t_x, t_y), c} \\
    s_{\avg}(t_x,t_y) &= \frac{1}{|C|} \sum_{c \in \conceptset}S_{(t_x, t_y), c} \\
    % s_{\corr}(t_x,t_y) &= \frac{\cov(S_{(t_x, t_y), c=1,..,|\conceptset|}, P_{(t_x, t_y), c=1,..,|\conceptset|})}{\sigma_{S_{(t_x, t_y), c=1,..,|\conceptset|}} \sigma_{P_{(t_x, t_y), c=1,..,|\conceptset|}}}
    s_{\corr}(t_x,t_y) &= \rho(S_{(t_x, t_y), c=1,..,|\conceptset|}, P_{(t_x, t_y), c=1,..,|\conceptset|})
\end{aligned}
\end{align}
where $\rho$ denotes Pearson correlation. 
\emph{$s_{\max}$}, \emph{$s_{\avg}$} $\!\in\!$ [0,$\infinity$) show separation between unnormalized class-histograms; and
\emph{$s_{\corr}$} $\in$ [-1, 1] shows agreement between $S$ and $P$.
We build $S_{\max}$, $S_{\avg}$ and $S_{\corr}$ by computing Equation~\eqref{eqn:stats} $\forall (t_x, t_y) \in \Omega$. 
Metrics' complementary may lead to relevant \emph{concepts} different to pathological understanding.
% \eg \emph{Average} metric may assist pathologists to discover other relevant \emph{concepts}.

% where $\cov$ and $\sigma$ define the covariance and standard deviation respectively.

\iffalse
Specifically, for each class-pair $t_x, t_y \in \tumorset$,
\begin{inparaenum}[(i)]
\item The maximum measures the maximum separability between $t_x$ and $t_y$ for any $c \in \conceptset$.
\item The expectation measures the average separability between $t_x$ and $t_y$ for all $c \in \conceptset$.
\item The correlation measures the correlation between concept-wise separability scores and prior pathological knowledge.
\end{inparaenum}
\fi

\textbf{Risk:}
We \emph{conceptually} introduce the notion of risk as a weight to indicate the cost of misclassifying a sample of class $t_x$, erroneously as class $t_y$ \cite{ThaiNghe2010, He2013}.
Indeed, misclassifying a malignant tumor as a benign tumor is riskier than misclassifying it as an atypical tumor. 
Thus, we construct a risk vector $R \in \mathbb{R}^{\Omega}$. In this work, each entry in $R$ defines the symmetric risk of differentiating $t_x$ from $t_y$ measured as the number of class-hops needed to evolve from $t_x$ to $t_y$.

% The weight of $(t_x, t_y) \in \tumorset$ differentiation indicates the risk of misclassifying a sample of class $t_x$, erroneously as class $t_y$.
% For clinical purposes, pair-wise differentiation tasks typically weigh unequally. 
% For instance, one can imagine that misclassifying a malignant tumor as a benign tumor is riskier than misclassifying it as an atypical tumor. 
% 1. high level idea 
% In this work, we include the notion of risk $R \in \mathbb{R}^{\Omega}$ as a \emph{conceptual element} for aggregating pair-wise class separability scores $\forall (t_x, t_y) \in \tumorset$ obtained using an explainer~\cite{ThaiNghe2010, He2013}.
% % 2. illustrate 
% Indeed, one can imagine that misclassifying a malignant tumor as a benign tumor is riskier than misclassifying it as an atypical tumor. 
% % 3. actual example eg KG
% We define risk  - disease progression. For simplicity, we keep the risk symmetric, \ie and measure it as number of hops one-hop distance.

\textbf{Metrics:}
Finally, we propose three quantitative metrics based on class separability to assess an explainer quality. The metrics are computed as the risk weighted sum of the statistics of separability scores, \ie, 
\begin{inparaenum}[(1)]
\item \emph{maximum separability} $S_{\max, R} := S_{\max} \odot R$,
\item \emph{average separability} $S_{\avg, R} := S_{\avg} \odot R$,
\item \emph{correlated separability} $S_{\corr, R} := S_{\corr} \odot R$, where $\odot$ defines the Hadamard product.
\end{inparaenum}
The first two metrics are pathologist-independent, and the third metric requires expert pathologists to impart the domain knowledge in the form of pathological prior $P$. Such prior can be defined individually by a pathologist or collectively by consensus of several pathologists, and it is independent of the algorithm generated explanations.

\begin{algorithm}[ht]
%\SetAlgoLined
\KwIn{$\dataset=\{(D_i^t, \importanceset_i^t) \}, t\in \tumorset, i \in N_t$ }
\Parameter{$\tumorset$, $\conceptset$, $\attributeset_c$, $\kset$}
\KwResult{$S \in \mathbb{R}^{{|\tumorset| \choose 2} \times |\conceptset|}$}  % can be removed 
\For(\tcp*[h]{go over concepts}){c in $\conceptset$ }{
 \For(\tcp*[h]{go over nuclei thresh}){k in $\kset$}{   
   \For(\tcp*[h]{go over attributes}){a in $\attributeset_c$}{
     \For(\tcp*[h]{go over classes}){t in $\tumorset$}{
    %   $\mbox{var} \gets \{D_i^t(a)|I_i^t\in\text{top}(k), \forall i \in N_t\}$\;
    %   $\mbox{var} \gets \{\text{sorted}_{I_i^t}(D_i^t(a))[:k]\}$\;
      $\mbox{var} \gets D_i^t(a)[:k]$  \tcp*[h]{sorted $I_i^t$}
      $H_t^{(k)}(a) \gets \mbox{histogram}(\mbox{var})$
    }
    \For(\tcp*[h]{go over class pairs}){$(t_x,t_y)$ in ${|\tumorset| \choose 2}$}{
       $s_a^{(k)}(t_x, t_y) \gets d(H_{t_x}^{(k)}(a), H_{t_y}^{(k)}(a))$ 
    }
   }
   $s_c^{(k)}(t_x, t_y) \gets \frac{1}{|\attributeset_c|} \sum_{a \in \attributeset_c} s_a^{(k)}(t_x, t_y)$ 
  }
  $S_{(t_x, t_y), c} \gets \mbox{AUC}_{k \in \kset} (s_c^{(k)}(t_x, t_y))$
 }
 \caption{Class separability computation.}
 \label{algorithm:metric}
\end{algorithm}
%

%
\iffalse
\begin{algorithm}[ht]
\SetAlgoLined
\KwIn{$\dataset=\{(D_i^t, \importanceset_i^t) \}, t\in \tumorset, i \in N_t$ }
\Parameter{$\tumorset$, $\conceptset$, $\attributeset_c$, $\kset$}
\KwResult{$S \in \mathbb{R}^{{|\tumorset| \choose 2} \times |\conceptset|}$}  % can be removed 
\For(\tcp*[h]{go over concepts}){c in $\conceptset$ }{
 \For(\tcp*[h]{go over attributes}){a in $\attributeset_c$}{
   \For(\tcp*[h]{go over nuclei thresh}){k in $\kset$}{   
     \For(\tcp*[h]{go over classes}){t in $\tumorset$}{
      $\mbox{var} \gets \{D_i^t(a)|I_i^t>\text{top}(k), \forall i \in N_t\}$\;
      $H_t^{(k)}(a) \gets \mbox{histogram}(\mbox{var})$\;
    }
    \For(\tcp*[h]{go over class pairs}){$(t_x,t_y)$ in ${|\tumorset| \choose 2}$}{
       $s_a^{(k)}(t_x, t_y) \gets d(H_{t_x}^{(k)}(a), H_{t_y}^{(k)}(a))$ 
    }
   }
   $S_{(t_x, t_y), a} \gets \mbox{AUC}_{k \in \kset} (s_a^{(k)}(t_x, t_y))$\;
  }
  $S_{(t_x, t_y), c} \gets \frac{1}{|\attributeset_c|} \sum_{a \in \attributeset_c} S_{(t_x, t_y), a}$ \;
 }
 \caption{Class separability computation.}
 \label{algorithm:metric}
\end{algorithm}
\fi
%

\section{Results}
This section describes the analysis of $\cg$ explainability for breast cancer subtyping. We evaluate three types of graph explainers and quantitatively analyze the explainer quality using the proposed class separability metrics.

\subsection{Dataset}
We experiment on BReAst Cancer Subtyping (BRACS), a large collection of breast tumor $\troi$s \cite{Pati2020}. BRACS consists of 4391 $\troi$s at 40$\times$ resolution from 325 H\&E stained breast carcinoma whole-slides.
% @TODO: remove to save space?
% The whole-slides are scanned with Aperio AT2 scanner at 0.25 $\mu$m/pixel for 40$\times$ resolution.
The $\troi$s are annotated by the consensus of three pathologists as, 
\begin{inparaenum}[(1)]
\item Benign (B): normal, benign and usual ductal hyperplasia,
\item Atypical (A): flat epithelial atypia and atypical ductal hyperplasia, and
\item Malignant (M): ductal carcinoma \emph{in situ} and invasive.
\end{inparaenum}
The $\troi$s consist of an average \#pixels=$3.9\pm4.3$ million, and average \#nuclei=$1468\pm1642$, and are stain normalized using \cite{Stanisavljevic2019}. The train, validation, and test splits are created at the whole-slide level, including 3163, 602, and 626 $\troi$s. 
% Further dataset statistics are provided in Appendix.

\begin{figure*}[!t]
\centering
\centerline{\includegraphics[width=0.81\linewidth]{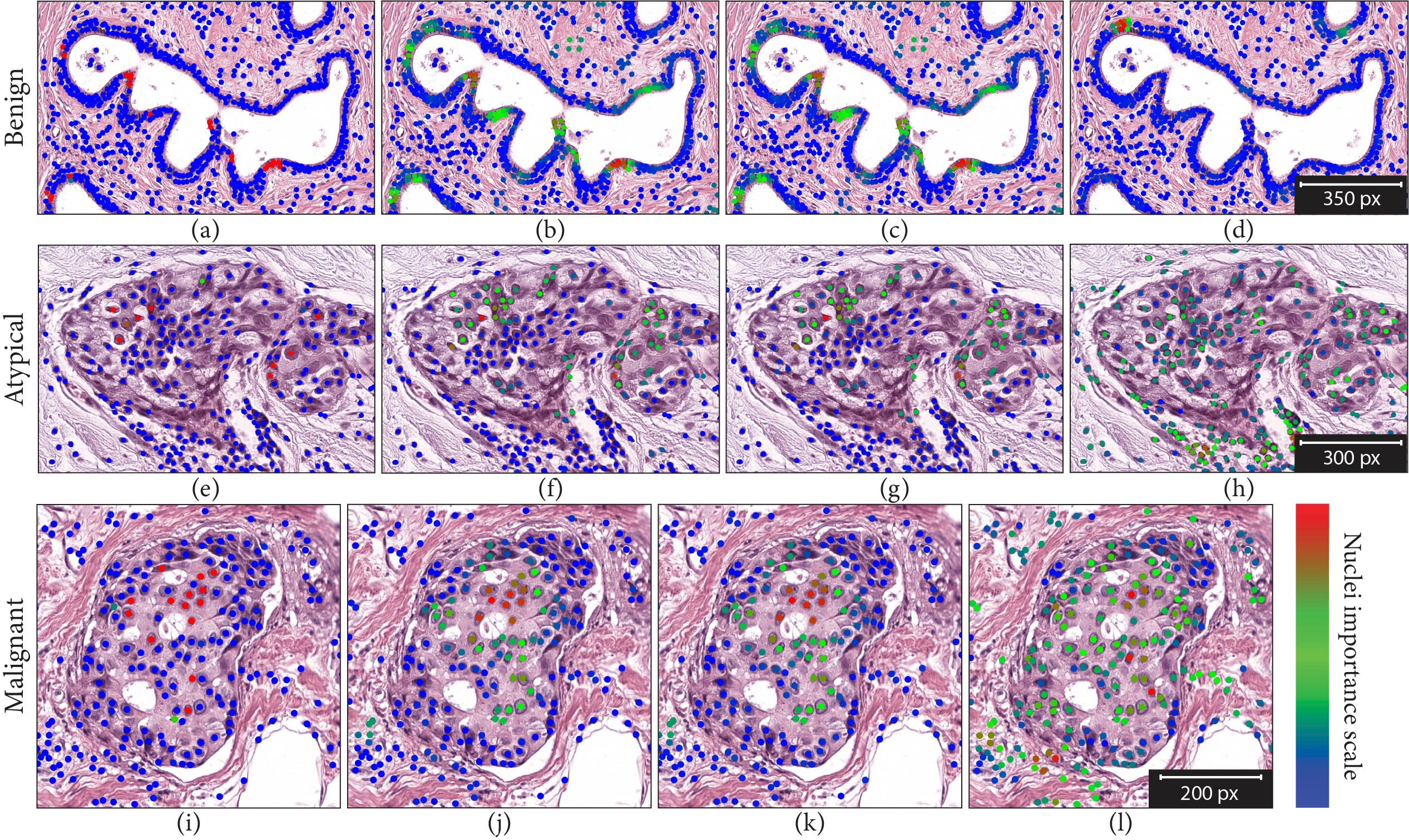}}
\caption{Qualitative results. The rows represent the cancer subtypes, \ie Benign, Atypical and Malignant, and the columns represent the graph explainability techniques, \ie $\gnnexplainer$, $\graphgradcam$, $\graphgradcampp$, and $\graphlrp$. Nuclei-level importance ranges from blue (the least important) to red (the most important).}
\label{fig:qualitative_results}
\end{figure*}

\subsection{Training}
% hardware/software 
% We conducted our experiments using PyTorch~\cite{Paszke2019} on NVIDIA Tesla P100 GPUs with POWER9 processors. All the experiments pertaining to graphs were implemented using the DGL library~\cite{Wang2019}.
We conducted our experiments using PyTorch~\cite{Paszke2019} and the Deep Graph Library (DGL)~\cite{Wang2019}.
The $\gnn$ architecture for $\cg$ classification is presented in Section~\ref{sec:graph_learning}. 
The $\cg$ classifier was trained for 100 epochs using Adam optimizer~\cite{Kingma2015}, $10^{-3}$ learning rate and 16 batch size.
% accuracy of the black-box GNN 
The best $\cg$-classifier achieved $74.2\%$ weighted F1-score on the test set for the three-class classification.
Average time for processing a 1K$\times$1K $\troi$ on a NVIDIA P100 GPU is 2s for $\cg$ generation and 0.01s for $\gnn$ inference.
% 0.3s for explanation generation and,
% 0.2 for the metric computation.

\subsection{Qualitative assessment}
% Introduce figure 3:
Figure~\ref{fig:qualitative_results} presents explanations, \ie nuclei importance maps, from four studied graph explainers.
% Observation 1: GradCAM is close to GradCAM++
We observe that $\graphgradcam$ and $\graphgradcampp$ produce similar importance maps. %The highlighted regions are spatially defined and correspond to areas with similar nuclei types. 
% Observation 2: GNNExplainer 
The $\gnnexplainer$ generates almost binarized nuclei importances. 
% Observation 3: consistency between gradient-based maps and GNNExplainer 
Interestingly, the gradient and pruning-based techniques consistently highlight similar regions.
% In Figure~\ref{fig:qualitative_results},
Indeed, the approaches focus on relevant epithelial region and unfocus on stromal nuclei and lymphocytes outside the glands.
% Observation 4: GraphLRP produces the most scattered maps without defined highlighted regions 
Differently, $\graphlrp$ produces less interpretable maps through high spatial localization (Figure~\ref{fig:qualitative_results}\textcolor{red}{(d)}) or less spatial localization (Figure~\ref{fig:qualitative_results}\textcolor{red}{(h,l)}).
% What are we missing with qualitative analysis ?
Qualitative visual assessment of Figure~\ref{fig:qualitative_results} conclude that, 
\begin{inparaenum}[(1)]
\item \emph{fidelity} preserving explainers result differently based on the underlying mechanism,
\item high \emph{fidelity} does not guarantee straightforward pathologist-understandable explanations,
\item qualitative assessment cannot rigorously compare explainers' quality, and
\item large-scale tedious pathological evaluation is inevitable to rank the explainers.
\end{inparaenum} 

% 1. not rigorous
%By restricting the analysis to qualitative considerations, we cannot rigorously rank the explainability methods.
% 2. would need time-consuming study with pathologists - 
%Furthermore, only a large-scale tedious sample-level analysis conducted with pathologist can define the relevance of an explanation. 

\iffalse
% A note on fidelity : MOVED TO INTRODUCTION
%All the considered algorithms have an internal mechanism to generate high-fidelity explanations, \ie a forward pass of the nuclei-weighted graph results in similar logit values as the original graph.
% For instance, $\gnnexplainer$ ensures fidelity by explicitly finding a mask such that the masked graph returns the same prediction as the original graph. The gradient-based and layer-wise relevance propagation approaches ensure fidelity by back-propagating the predicted logit value till the input.
\fi

% Table 1: Max, Exp & correlation scores
\begin{table*}[t]
\centering

\begin{tabular}{l|c|ccc|c|c|c|c}
  \hline
  \multicolumn{2}{l|}{Tasks $(\Omega)$}  & B vs. A & B vs. M & A vs. M & \multicolumn{4}{c}{B vs. A vs. M} \\
  \hline
  \multicolumn{2}{l|}{Accuracy (in \%)} & $77.19$ &	$90.29$ &  $80.42$ & \multicolumn{4}{c}{$74.92$} \\
  \hline
  
  Explainer & \multicolumn{4}{c|}{Metric $\;\;\forall \; (t_x,t_y) \in \Omega$ ($\uparrow$)} & \multicolumn{2}{c|}{Agg. Metric w/o Risk ($\uparrow$)} & \multicolumn{2}{c}{Agg. Metric w/ Risk ($\uparrow$)} \\
  
  \hline
  $\gnnexplainer$ & \parbox[t]{2mm}{\multirow{5}{*}[-0.4ex]{\rotatebox[origin=c]{90}{$s_{\max}(t_x, t_y)$}}}
   & $\mathbf{3.26}$ &	$\mathbf{6.24}$ &	$\mathbf{3.48}$ &
  \parbox[t]{2mm}{\multirow{5}{*}[-0.4ex]{\rotatebox[origin=c]{90}{$S_{\max}$}}}
  & $\mathbf{12.98}$ &
  \parbox[t]{2mm}{\multirow{5}{*}[-0.4ex]{\rotatebox[origin=c]{90}{$S_{\max,R}$}}}
  & $\mathbf{19.22}$  \\ 
  $\graphgradcam$ & & $1.24$ &	$4.41$ &	$3.36$ & & $9.01$ & & $13.42$  \\ 
  $\graphgradcampp$ & & $1.27$ &	$\underline{4.42}$ &	$\underline{3.40}$ & & $\underline{9.09}$ & & $\underline{13.51}$  \\ 
  $\graphlrp$ & & $\underline{2.33}$ &	$2.46$ &	$1.28$ & & $6.07$ & & $8.53$  \\ 
  $\random$ & & $1.02$ &	$1.26$ &	$1.11$ & & $3.39$ & & $4.65$  \\ 
  
  \hline
  $\gnnexplainer$ & \parbox[t]{2mm}{\multirow{5}{*}[-0.4ex]{\rotatebox[origin=c]{90}{$s_{\avg}(t_x, t_y)$}}}
  & $\mathbf{1.54}$ &	$\mathbf{2.78}$ &	$1.93$          &
  \parbox[t]{2mm}{\multirow{5}{*}[-0.4ex]{\rotatebox[origin=c]{90}{$S_{\avg}$}}}
  & $\mathbf{6.25}$ &
  \parbox[t]{2mm}{\multirow{5}{*}[-0.4ex]{\rotatebox[origin=c]{90}{$S_{\avg,R}$}}}
  & $\mathbf{9.03}$  \\ 
  $\graphgradcam$ &   & $1.15$          &	$2.57$          &	$\underline{2.08}$ &         & $5.80$  &        & $8.37$  \\ 
  $\graphgradcampp$ &  & $1.18$          &	$\underline{2.58}$          &	$\mathbf{2.09}$& & $\underline{5.85}$  &        & $\underline{8.43}$  \\ 
  $\graphlrp$  &      & $\underline{1.38}$          &	$1.59$          &	$1.47$ &         & $4.44$  &        & $6.03$  \\ 
  $\random$  &     & $1.05$          &	$1.00$          &	$0.95$ &         & $3.00$  &        & $4.00$  \\ 
  
  \hline
  $\gnnexplainer$ & \parbox[t]{2mm}{\multirow{5}{*}[-0.4ex]{\rotatebox[origin=c]{90}{$s_{\corr}(t_x, t_y)$}}}
  & $-0.02$ &	$0.36$          &	$0.38$  &   
  \parbox[t]{2mm}{\multirow{5}{*}[-0.4ex]{\rotatebox[origin=c]{90}{$S_{\corr}$}}}
  & $0.72$  &
  \parbox[t]{2mm}{\multirow{5}{*}[-0.4ex]{\rotatebox[origin=c]{90}{$S_{\corr,R}$}}}
  & $1.08$  \\ 
  $\graphgradcam$ &  & $\underline{-0.01}$          &	$\underline{0.57}$          &	$\underline{0.58}$   &       & $\underline{1.14}$   &       & $\underline{1.71}$  \\ 
  $\graphgradcampp$ & & $\mathbf{-0.01}$          &	$\mathbf{0.58}$ &	$\mathbf{0.59}$ & & $\mathbf{1.16}$ & & $\mathbf{1.74}$  \\ 
  $\graphlrp$ &      & $-0.15$          &	$-0.49$         &	$-0.23$  &       & $-0.87$   &      & $-1.36$  \\ 
  $\random$   &      & $-0.37$           &	$-0.31$         &	$-0.18$  &       & $-0.86$   &      & $-1.17$  \\ 
  
  \hline
\end{tabular}
\caption{Quantitative assessment of graph explainers: $\gnnexplainer$, $\graphgradcam$, $\graphgradcampp$ and $\graphlrp$, using proposed \emph{maximum, average}, and \emph{correlated separability} metrics. Results are provided for each pair-wise breast subtyping tasks, and are aggregated w/o and w/ risk weighting, \ie $S_{\max}$ and $S_{\max,R}$. The first and second best values are indicated in $\textbf{bold}$ and $\underline{\mbox{underline}}$.}
\label{tab:table1}
\end{table*}

% Table 2:
\begin{table*}
  \centering
  \begin{tabular}{l|ccc|cc}
    \hline
    Concept \scriptsize{(Attributes)} \normalsize{/ Tasks ($\Omega$)} & B vs. A & B vs. M & A vs. M & w/o risk ($\uparrow$) & w/ risk ($\uparrow$) \\
    \hline
    Size   \scriptsize{(area)}         & $\mathbf{3.26}$ &	$\mathbf{6.24}$ &	$\mathbf{3.47}$ & $\mathbf{12.97}$ & $\mathbf{19.21}$  \\ 
    Shape  \scriptsize{(perimeter, roughness, eccentricity, circularity)}        & $1.27$ &	$2.23$ &	$1.60$ & $5.10$ & $7.34$  \\ 
    Shape variation \scriptsize{(shape factor)} & $0.69$ &	$2.30$ &	$1.99$ & $4.97$ & $7.28$  \\ 
    Density  \scriptsize{(mean density, std density)}      & $1.01$ &	$0.80$ &	$0.52$ & $2.33$ & $3.14$  \\ 
    Chromaticity \scriptsize{(GLCM contrast, homogeneity, ASM, entropy, variance)}      & $\underline{1.44}$ &	$\underline{2.31}$ &	$\underline{2.07}$ & $\underline{5.82}$ & $\underline{8.13}$  \\ 
    \hline
    \emph{Average separability} ($\uparrow$)  & $1.54$ &	$2.78$	& $1.93$ & $6.25$ & $9.03$  \\
    \hline
  \end{tabular}
  \caption{Quantification of \emph{concepts} for pair-wise and aggregated class separability in $\gnnexplainer$. The first and second best values are indicated in $\textbf{bold}$ and $\underline{\mbox{underline}}$. The per-\emph{concept} \emph{attributes} are presented in the first column.}
  \label{tab:table2}
\end{table*}

\subsection{Quantitative results}

For cancer subtyping, relevant \emph{concepts} are nuclear morphology and topology \cite{Rajbongshi2018,Kashyap2018,Nguyen2017,Allison2016}. Here, we focus on nuclear morphology, \ie $\conceptset$ = \{\emph{size}, \emph{shape}, \emph{shape variation}, \emph{density}, \emph{chromaticity}\}.
Table~\ref{tab:table2} lists the \emph{attributes} $\attributeset_c, \forall c \in \conceptset$. In our experiments, we select $\kset = \{5, 10, ..., 50 \}$ nuclei per $\cg$.
% introduce the random baseline
We further introduce a $\random$ explainer via \emph{random} nuclei selection strategy per $\cg$ to assess a lower bound per quantitative metric.
% introduction to Table 1
Table~\ref{tab:table1} presents the statistics of pair-wise class separability and aggregated separability w/ and w/o risk to assess the studied explainers quantitatively. Also, for each class pair $(t_x, t_y)$, we compute classification accuracy by using the $\cg$s of type $t_x$, $t_y$.
% $\forall \; \mbox{\troi} \in t_x, t_y$. 

% Analyze max and average
Noticeably, $\gnnexplainer$ achieves the best \emph{maximum} and \emph{average separability} for majority of pair-wise classes. $\graphgradcampp$ and $\graphgradcam$ followed $\gnnexplainer$ except for (B vs. A), where $\graphlrp$ outperforms them. All explainers outperform $\random$ which conveys that the quality of the explainers' explanations are better than random.
Notably, $\graphgradcam$ and $\graphgradcampp$ quantitatively perform very similarly, which is consistent with our qualitative analysis in Figure \ref{fig:qualitative_results}.
% \pus{Intuitively, spatial contribution (size of a graph) towards graph explainability is less significant.}
%
% Analyze max and average behavior w.r.t. performance accuracy
Interestingly, a positive correlation is observed between pair-wise class accuracies and \emph{average separability} for the explainers, \ie better classification leads to better \emph{concept} separability, and thus produces better explanations. Further, the observation does not hold for $\random$ generated explanations, which possesses undifferentiable \emph{average concept} separability. %The findings convey that better classification leads to better explanations for an explainer.
%\pus{It conveys that black-box $\gnn$ performance is positively correlated with pathologically relevant \emph{concepts}.}
%It signifies that the observation is valid only for selecting relevant nuclei. This hints that a better black-box $\gnn$ performance is positively correlated with pathologically relevant \emph{concepts}.

%Indeed, high model performance, \eg B vs. M, correlates with high \emph{average separability}. This observation does not hold for the $\random$ baseline where the \emph{average separability} is not a good predictor of the per-class pair model performance. This means that the connection between model performance and \emph{concept} values applies only when selecting relevant nuclei. This is a first hint that the pathologically relevant \emph{concepts} aligns with the black box reasoning of the GNN, \ie when the model performance is good, then we are able to separate the concepts. 

% GradCAM and GradCAM++ perform similarly, as suggested by our qualitative analysis.
%\graphgradcam$ and $\graphgradcampp$ have similar \emph{maximum} and \emph{average separability} performance. This observation is consistent with our qualitative analysis that highlighted similar node importance for those two approaches. 
% LRP matching qualitative results is the worst performer 
%$\graphlrp$ is outperformed by the three other explainers while remaining overall better than the $\random$ baseline. 
% comment on the relation between class-pair-wise model performance and the average separability scores 

% Analyze correlation
To obtain pathological prior to compute \emph{correlation separability}, we consulted three pathologists to rank the \emph{concepts} in terms of their relevance for discriminating each pair of classes. For instance, given an atypical $\troi$, we asked how important is nuclear \emph{shape} to classify the $\troi$ as \emph{not} benign and \emph{not} malignant.
Acquired \emph{concept} ranks for each class pair are \emph{min-max} normalized to output prior matrix $P$.
%To this end, a dataset of 100 $\troi$s per class is used. The ranked \emph{concept} are averaged across $\troi$s belonging to a pair of classes, followed by a \emph{min-max} normalization across all \emph{concepts}. The outcome is a normalized prior matrix $P \in \mathbb{R}^{\Omega \times |\conceptset|}$.
%
We observe that $\gnnexplainer$, $\graphgradcam$ and $\graphgradcampp$ have positive \emph{correlated separability} for (B vs. M), (A vs. M), and nearly zero values for (B vs. A). It shows that the explanations for (B vs. M) and (A vs. M) bear similar relevance of \emph{concepts} as the pathologists, and focus on a different relevance of \emph{concepts} for (B vs. A). 
$\graphgradcampp$ has the best overall agreement at the \emph{concept}-level with the pathologists, followed by $\graphgradcam$ and $\gnnexplainer$. $\random$ agrees significantly worse than the three explainers, and $\graphlrp$ has the least agreement.
%
% comment on the correlation: high-level introduction / how we got it for this work / comment the actual scores 
%The \emph{correlation separability} was computed by looking at the correlation between the separability matrix $S$ and the prior pathological knowledge $P$. The prior can be obtained in different ways. In this paper, we asked pathologists to assign an importance score to each \emph{concept} to discriminate each class pair. For instance, given an atypical sample, we ask how important is the \emph{concept} shape for concluding that this sample is \emph{not} malignant. The outcome is a normalized prior matrix $P \in \mathbb{R}^{{|\tumorset| \choose 2} \times \conceptset}$, where each importance ranges from zero to one.
% 
% The \emph{expectation separability} also provides a way to rank the task complexity. 
% For instance, when studying B vs A, arguably the most complex task, the \emph{expectation separability} is minimum. 
% This observation is also consistent with the correlation scores that are the highest 
% @PUS: do we want to conclude that when we are good at performing the task (from confusion matrix), then we have high max and average separability, and that it empirically matches with good correlation separability with prior. Interesting point: for random the average separability scores are NOT predictors of the model performance! Which means that the connection between model performance and concept values apply only to relevant nuclei! 
%
% Comments on Table 2
Table~\ref{tab:table2} provides more insights by highlighting the per-\emph{concept} metrics of $\gnnexplainer$. Nuclear \emph{size} is the most relevant \emph{concept}, followed by \emph{chromaticity} and \emph{shape variation}. Comparatively nuclear \emph{density} is the least relevant \emph{concept}.

%We observe large differences between the \emph{concept} scores. For instance, the selected nuclei's nuclei size is a highly discriminative sample, while the nuclei spacing, \ie nuclei density, is less informative.

% TODO: PUS
% 11. link it to the idea that those 2 scores define how important a concept is given an interpretability method & how good is an interpretability method given a concept or a set of concept. 

% 12. introduce the correlated separability as a means to measure the extend to which a set of concept importance align with pathological prior in order to distinguish between pair of classes 

\section{Conclusion}
%you can shorten conclusion a bit
% claims: 1. pixel-wise analysis to entity-graph
In this work, we presented an approach for explaining black-box $\deeplearning$ solutions in computational pathology. We advocated for biological entity-based analysis instead of conventional pixel-wise analysis, thus providing an intuitive space for pathological understanding.
% claims: 2. exhaustive graph explainers
We employed four graph explainability techniques, \ie graph pruning ($\gnnexplainer$), gradient-based saliency ($\graphgradcam$, $\graphgradcampp$) and
layerwise relevance propagation ($\graphlrp$), to explain ``black-box" $\gnn$s processing the entity graphs. 
% We employed four graph explainability techniques to explain the ``black-box" $\gnn$s processing of the entity graphs. 
% claims: 3. novel set of quantitative metrics 
We proposed a novel set of user-independent quantitative metrics expressing pathologically-understandable \emph{concepts} to evaluate the graph explainers, which relaxes the exhaustive qualitative assessment by expert pathologists. 
%
% outcome: 1. the nuclei selection strategy of the explainers is not random and is aligned with pathological prior 
% The quantitative assessment
Our analysis concludes that the explainer bearing the best class separability in terms of \emph{concepts} is $\gnnexplainer$, followed by $\graphgradcampp$ and $\graphgradcam$. $\graphlrp$ is the worst explainer in this category while outperforming a randomly created explanation.
%
% outcome: 2. the pair-wise separability scores are predictors of the pair-wise model performance 
We observed that the explainer quality is directly proportional to the $\gnn$'s classification performance for a pair of classes.
%
% outcome: 3. correlation metrics, highlights relevant concepts
Furthermore, $\graphgradcampp$ produces explanations that best agrees with the pathologists in terms of \emph{concept} relevance, and objectively highlights the relevant set of \emph{concepts}.
Considering the expansion of entity graph-based processing, such as radiology, computation biology, satellite and natural images, graph explainability and their quantitative evaluation is crucial.
The proposed method encompassing domain-specific user-understandable terminologies can potentially be of great use in this direction. It is a meta-method that is applicable to other domains and tasks by incorporating relevant entities and corresponding \emph{concepts}.
For instance, with entity-graph nodes denoting car/body parts in Stanford Cars~\cite{krause2013}/ Human poses~\cite{andriluka2014}, and expert knowledge available on car-model/ activity, our method can infer relevant parts by quantifying
% agreement of these with experts.
their agreement with experts.

% \newpage
{\small
\bibliographystyle{ieee_fullname}
\bibliography{main}
}

\clearpage
\section*{Post-hoc explainers}
In this section, we present the details of the considered graph explainability techniques (explainers) in this work: $\graphlrp$, $\graphgradcam$, $\graphgradcampp$, $\gnnexplainer$, and $\random$.

\subsection*{Notation}
We define an attributed undirected entity graph $G:= (V, E, H)$ as a set of nodes $V$, edges $E$, and node attributes $H \in \mathbb{R}^{|V| \times d}$. $d$ denotes the number of attributes per node, and $|.|$ denotes set cardinality.
% add what is an edge 
We denote an edge between nodes $u$ and $v$ as $e_{uv} \in E$.
% add what is an adjacency matrix
The graph topology is defined by a symmetric graph adjacency, $A \in \mathbb{R}^{|V| \times |V|}$, where $A_{uv} = 1$ if $e_{uv} \in E$.
$H_{n,k}$ expresses the $k$-th attribute of the $n$-th node. 
The forward prediction of a cell-graph $G_{\cg}$ is denoted as, $y = \model(G_{\cg})$, where $\model$ is a pre-trained $\gnn$, and $y \in \mathbb{R}^{|\tumorset|}$ are the output logits.
Notation $y(t), \; t \in \tumorset$ denotes the output logit of the $t$-th class.
We refer to the logit of the predicted class as $y_{\max} = \max_{t \in \tumorset} y(t)$, and the predicted class as $t_{\max} = \argmax_{t \in \tumorset} y(t)$.

\subsection*{Layerwise relevance propagation: $\graphlrp$} \label{sec:graphlrp}

% introduction to LRP 
Layerwise Relevance Propagation (LRP)~\cite{Bach2015} is a feature attribution based post-hoc explainer.
LRP explains an output logit by determining the individual contribution of each input element to the logit value. An output logit, defined as the output \emph{relevance} for a given class, is layerwise back-propagated until the input to compute the positive or negative impact of the input elements on the output logit.
% detailed description 
LRP, initially proposed for fully connected layers ($\lrpfc$), works as follows. Given a pre-trained fully connected layer $W \in \mathbb{R}^{z_1 \times z_2}$ between layer $1$ and layer $2$, where $z_1$ and $z_2$ are the number of neurons in layer $1$ and layer $2$, respectively, we compute the contributions of a neuron $i, \; i \in \{1, ..., z_1\}$ using the propagation rules in~\cite{Montavon2015}. In this work, we are interested in identifying the input elements \emph{positively} contributing to the prediction. To this end, we use the $z^+$ propagation rule that back-propagates the \emph{positive} neuron contribution from layer $2$ to layer $1$ as:
\begin{align}
\tag{$\lrpfc$}\label{eq:lrp_fc}
    R_i = \sum_j^{z_2} \frac{f_i |w_{ij}|}{\sum_k^{z_1} f_k |w_{kj}|} R_j
\end{align}
%
% where, $d^{(l)}$ and $d^{(l+1)}$ are the number of neurons in layer $l$ and $l+1$ respectively.
where $|w_{ij}|$ is the absolute value of the weight between $i$-th and $j$-th neuron in layer $1$ and $2$, respectively. $f_i$ denotes the activation of the $i$-th neuron in layer $l$.

The extension from $\lrpfc$ to LRP for graph isomorphism network (GIN) layers ($\graphlrp$) is achieved by following the observations in \cite{Schwarzenberg2019}. First, the \emph{aggregate step} in $\gnn$ corresponds to projecting the graph's adjacency matrix on the node attribute space. For simplicity, assuming a 1-layer MLP as an update function, the GIN layer with \emph{mean} aggregator can be re-written in its global form as:
\begin{align}
    H^{(l+1)} = \sigma \Big( W^{(l)} (I+\Tilde{A}) H^{(l)} \Big)
\end{align}
where $\Tilde{A}$ is the degree-normalized graph adjacency matrix, \ie $\Tilde{A}_{ij} = \frac{1}{|\mathcal{N}(i)|} A_{ij}$. $\sigma$ is the $\mbox{ReLU}$ activation function. Second, this representation allows us to treat the term $(I + \Tilde{A})$ as a regular, fully connected layer. We can then apply the $z^+$ propagation rule with weights $w_{ij}$ defined as:
\begin{align}
w_{ij} &= 1 \quad \text{if} \; i=j\\
w_{ij} &= \frac{1}{|\mathcal{N}(i)|} \quad \text{if} \; e_{ij} \in E \\
w_{ij} &= 0 \quad \text{otherwise}
\end{align}
LRP outputs an importance score for each node $i$ in the input graph.
% Thus, the propagation rule for LRP-GIN with 1-layer MLP becomes:
% %
% \begin{align}\tag{$\graphlrp$}\label{eq:lrp_gin}
%     R_i = \sum_m^{|V|} \frac{h_i^{(l)} e_{im}}{\sum_{n \in \textbf{1}_{e_{nm}}} h_n^{(l)}e_{nm}}  \Big( \sum_j^{d^{(l+1)}} \frac{f_i^{(l)} |w_{ij}^{(l)}|}{\sum_k^{d^{(l)}} f_k^{(l)} |w_{kj}^{(l)}|} R_j \Big )
% \end{align}
% %
% where $e_{im} \in (I + \Tilde{A})$, and $\mathbf{1}_{e_{nm}}$ is an indicator function returning one if $e_{nm} > 0$, and zero otherwise. \\

\subsection*{Saliency-based: $\graphgradcam$} \label{sec:gradcam}
% high-level introduction:
Grad-CAM~\cite{Selvaraju2017} is a feature attribution post-hoc explainer that identifies salient regions of the input driving the neural network prediction. It assigns importance to each element of the input to produce Class Activation Map~\cite{Zhou2016}. While originally developed for explaining $\cnn$s operating on images, $\gradcam$ can be extended to $\gnn$s operating on graphs~\cite{Pope2019}.

% more details 
$\graphgradcam$ processes in two steps. First, it assigns an importance score to each channel of a graph convolutional layer. The importance of channel $k$ in layer $l$ is computed by looking at the gradient of the predicted output logit $y_{\max}$ w.r.t. the node attributes at layer $l$ of the $\gnn$. Formally it is expressed as:
\begin{align}\label{eq:gradcam_weights}
    w^{(l)}_k = \frac{1}{|V|} \sum_{n=1}^{|V|} \frac{\partial y_{\max}}{\partial H^{(l)}_{n, k}}
\end{align}
In the second step, a node-wise importance score is computed using the forward node feature activations $H^{(l)}$ as:
\begin{align}\tag{$\graphgradcam$}\label{eq:graph_graph_cam}
    L(l, v) = \mbox{ReLU} \Big( \sum_k^{d^{(l)}} w^{(l)}_k H^{(l)}_{n, k} \Big)
\end{align}
where $L(l, v)$ denotes the importance of node $v \in V$ in layer $l$, and $d^{(l)}$ denotes the number of node attributes at layer $l$. 
Since we are only interested in the positive node contributions, \ie nodes that positively influence the class prediction, we apply a ReLU activation to the node importances.
Following prior work~\cite{Pope2019}, we take the average node importance scores obtained over all the $\gnn$ layers $l \in \{1, ..., L\}$ to obtain smoother node importance scores.

\subsection*{Saliency-based: $\graphgradcampp$} \label{sec:gradcampp}
$\graphgradcampp$ extends $\gradcampp$ \cite{Chattopadhay2018} to graph structured data. It improves the node importance localization by introducing node-wise contributions to channel importance scoring in Equation \ref{eq:gradcam_weights}. Specifically, the modification is presented as,
\begin{align}\label{eq:gradcampp_weights}
    w^{(l)}_k = \frac{1}{|V|} \sum_{n=1}^{|V|} \alpha_{n,k}^{(l)} \frac{\partial y_{max}}{\partial H^{(l)}_{n, k}}
\end{align}
where $\alpha_{n,k}^{(l)}$ are node-wise weights expressed for each attribute $k$ at layer $l$.
The derivation of a closed-form solution for $\alpha_{n,k}^{(l)}$ is analogous to the derivation in~\cite{Chattopadhay2018}, where the size of graph, \ie number of nodes, replaces the spatial dimensions of a channel as:
\begin{align}\label{eq:gradcampp_weights}
    \alpha_{n,k}^{(l)} = \frac{\frac{\partial^2 y_{max}}{(\partial H^{(l)}_{n, k})^2}}{2 \frac{\partial^2 y_{max}}{(\partial H^{(l)}_{n, k})^2} + \sum_{n=1}^{|V|} H^{(l)}_{n, k}\big(\frac{\partial^3 y_{max}}{(\partial H^{(l)}_{n, k})^3}\big)}
\end{align}
The subsequent node importance computation in $\graphgradcampp$ is same as $\graphgradcam$.

\subsection*{Graph pruning: $\gnnexplainer$} \label{sec:gnnexplainer}
The $\gnnexplainer$~\cite{Ying2019, Jaume2020} is a graph pruning based post-hoc explainer for explaining $\gnn$s. $\gnnexplainer$ is model-agnostic, \ie it can be used with any flavor of $\gnn$. Intuitively, $\gnnexplainer$ tries to find the minimum sub-graph $G_s \subset G$ such that the model prediction $y=\model(G)$ is retained. The inferred sub-graph $G_s$ is then regarded as the \emph{explanation} for $G$. This approach can be seen as a feature attribution method with \emph{binarized} node importance scores, \ie a node $v \in V$ has importance one if $v \in V_s$, and zero otherwise. 
Exhaustively searching $G_s$ in the space created by nodes $V$ and edges $E$ is infeasible due to the combinatorial nature of the task. Instead, $\gnnexplainer$ formulates the task as an optimization problem that learns a mask to activate or deactivate parts of the graph.
The initial formulation by~\cite{Ying2019}, developed for explaining node classification tasks, learns a mask over the edges, \ie over the adjacency matrix. 
Instead, we follow the prior work in~\cite{Jaume2020} to learn a mask over the nodes. Indeed, as we are concerned with classifying $G$, the optimal explanation $G_s$ can be a disconnected graph. Furthermore, in cell graphs, the nodes representing biological entities are more intuitive and substantial for disease diagnosis than edges, that are heuristically-defined. 

Formally, we seek to learn a mask $M_V$ such that the induced masked sub-graph $G_s$,
\begin{inparaenum}[(1)]
\item is as small as possible,
\item outputs a binary node importance, and
\item provides the same prediction as the original graph. 
\end{inparaenum} 
These constraints can be modeled by considering a loss function as:
\begin{align}\label{eq:explainer4}
    \loss = \loss_\kd(\hat{y}, y^{(m)}) + \alpha_{M_V} \sum_{i}^{|V|} \sigma(M_{V_i}^{(m)}) + \alpha_{\entropy} \entropy^{e}(\sigma(M_V^{(m)}))
\end{align}
where, $m$ is the optimization step and $\sigma$ is the sigmoid activation function.
The first term is a knowledge-distillation loss $\loss_\kd$ between  $\hat{y}=\model(G)$ and $y^{(m)}=\model(G_s)$ ensuring that $y^{(m)} \approx \hat{y}$. The second term aims to minimize the size of the mask $M_V$. The third term binarizes the mask by minimizing the element-wise entropy $\entropy^{e}$ of $M_V$.
Following previous work~\cite{Hinton2015}, $\loss_\kd$ is built as a combination of distillation and cross-entropy loss,
\begin{align} \label{eq:explainer5}
    \loss_\kd = \lambda \loss_\ce + (1 - \lambda) \loss_\dist \; \text{where} \: \lambda = \frac{\entropy^{e}(y^{(m)})}{\entropy^{e}(\hat{y})}
\end{align}
where $\loss_\ce$ is the regular cross-entropy loss and $\loss_\dist$ is the distillation loss.
When the element-wise entropy $\entropy^{e}(y^{(m)})$ increases, the term $\loss_\ce$ gets larger and reduces the probability of changing the prediction.
% add hyper parameters used
Each term in Equation~\ref{eq:explainer4} is empirically weighed such that their contributions to $\loss$ are comparable.
We set $\alpha_{M_V} = 0.005$ and $\alpha_{\entropy} = 0.1$. We learn $M_V$ using Adam optimizer with a learning rate of $0.01$.
$\loss$ is optimized for $1000$ steps with an early stopping mechanism, which triggers if the class prediction using $G_s$ is changed. Therefore, $G_s$ and $G$ always predict the same class, \ie $t_{\max}^{(m)} = \hat{t}_{\max} \; \forall m$.

% Table 4: TRoI dataset Statistics 
\begin{table*}[t]
\centering
\begin{tabular}{c|lccc|c}
  \toprule
  & Metric & Benign & Atypical & Malignant & Total \\
  \midrule
  \parbox[t]{2mm}{\multirow{3}{*}[0ex]{\rotatebox[origin=c]{90}{Image}}} & Number of images & 1741 & 1351 & 1299 & 4391 \\ 
  
  & Number of pixels (in million) & 3.9$\pm$3.54 & 1.62$\pm$1.48 & 6.35$\pm$5.2 & 3.9$\pm$4.3 \\
  
  & Max/Min pixel ratio & 180.1 & 75.3 & 128.6 & 235.6 \\ 
  
  \midrule
  \parbox[t]{2mm}{\multirow{3}{*}[0ex]{\rotatebox[origin=c]{90}{$\cg$}}} & Number of nodes & 1331$\pm$1134 & 635$\pm$510 & 2521$\pm$1934 & 1468$\pm$1642   \\ 
  
  & Number of edges & 4674$\pm$4131 & 2309$\pm$2110 & 8591$\pm$7646 & 5102$\pm$6089 \\ 
  
  & Max/Min node ratio & 312.5 & 416.7 & 312.5 & 434.8 \\ 
  
  \midrule
  \parbox[t]{2mm}{\multirow{3}{*}[0ex]{\rotatebox[origin=c]{90}{Image split}}} & Train & 1231 & 1008 & 928 & 3163 \\ [0.1cm]
  & Validation & 261 & 162 & 179 & 602 \\ [0.1cm]
  & Test & 249 & 185 & 192 & 626 \\ [0.1cm]
  \bottomrule
\end{tabular}
\caption{Statistics of BRACS dataset.}
\label{tab:dataset_statistics}
\end{table*}

\subsection*{Random selection: $\random$}
\label{sec:random}
The $\random$ baseline is implemented using a \emph{random} nuclei selection. The number of selected nuclei per $\troi$ is given by the threshold value $k \in \kset$.

\section*{BRACS dataset}
In this paper, the BRACS dataset is used to analyze $\cg$ explainability for breast cancer subtyping. The pixel-level and entity-level statistics of the dataset are presented in Table~\ref{tab:dataset_statistics}. Training, validation, and test splits are created at the whole-slide level for conducting the experiments. The details of the class-wise distribution of images in each split are presented in Table~\ref{tab:dataset_statistics}.

% Table 4: Concept extraction
\begin{table*}[t]
\centering
\begin{tabular}{c|l|c||l|l|l}
  \toprule
  Concept ($\conceptset$) & Attribute ($\attributeset$) & Computation & Benign & Atypical & Malignant \\
  \midrule
  % Size 
  Size & Area & $A(x)$ & Small & Small-Medium  & Medium-Large\\
  \midrule
  % shape
  \multirow{4}{*}{Shape} & Perimeter    & $P(x)$ & \multirow{4}{*}{Smooth} & \multirow{4}{*}{Mild irregular} & \multirow{4}{*}{Irregular}\\ [0.5em]
  & Roughness  & $\frac{P_{\text{ConvHull}}(x)}{P(x)}$  &  &  &  \\ [0.5em]
  & Eccentricity & $\frac{a_{\text{minor}}(x)}{a_{\text{major}}(x)}$  & & & \\ [0.5em]
  & Circularity  & $\frac{4 \pi A(x))}{P(x)^2}$ & & & \\
  \midrule
  % shape variation 
  Shape & \multirow{2}{*}{Shape factor} & \multirow{2}{*}{$\frac{4 \pi A(x)}{P_{\text{ConvHull}}^2}$} & \multirow{2}{*}{Monomorphic} & \multirow{2}{*}{Monomorphic} & \multirow{2}{*}{Pleomorphic}\\
  variation & & & & & \\
  \midrule
  % spacing 
  \multirow{2}{*}{Spacing} & Mean spacing & $\text{mean}(d_y | y \in \text{kNN(x)})$ & \multirow{2}{*}{Evenly crowded} & \multirow{2}{*}{Evenly spaced} & \multirow{2}{*}{Variable} \\ [0.5em]
  & Std spacing  & $\text{std}(d_y | y \in \text{kNN(x)})$ & & &\\
  \midrule
  % chromatin
  \multirow{7}{*}{Chromatin} & GLCM dissimilarity & $\sum_i\sum_j |i-j|p(i,j)$ & \multirow{7}{*}{Light} & \multirow{7}{*}{Hyperchromatic} & \multirow{7}{*}{Vesicular} \\ [0.5em]
  & GLCM contrast & $\sum_i \sum_j (i-j)^2 p(i,j)$ & \multirow{6}{*}{euchromatic} & & \\ [0.5em]
  & GLCM homogenity & $\sum_i\sum_j \frac{p(i,j)}{1 + (i-j)^2}$ & & & \\ [0.5em]
  & GLCM ASM & $\sum_i\sum_j p(i,j)^2$ & & & \\ [0.5em]
  & GLCM entropy & $-\sum_i\sum_j p(i,j)\log(p(i,j))$ & & & \\ [0.5em]
  & GLCM variance & $\sum_i\sum_j (i-\mu_i)^2 p(i,j)$ & & & \\ 
  & &  with $\mu_i = \sum_i \sum_j ip(i,j)$ & & & \\
  \bottomrule
\end{tabular}
\caption{Pathologically-understandable nuclear \emph{concepts}, corresponding measurable \emph{attributes}, and computations are shown in Columns 1, 2, 3, respectively. The expected \emph{concept} behavior for three breast cancer subtypes is shown in Columns 4, 5, 6, respectively.}
\label{tab:concept}
\end{table*}

\section*{Concepts and Attributes}
In this paper, we focus on pathologically-understandable nuclear \emph{concepts} $\conceptset$ pertaining to nuclear morphology for breast cancer subtyping. To quantify each $c \in \conceptset$, we use several measurable \emph{attributes} $\attributeset_c$. Table~\ref{tab:concept} presents the list of \emph{concepts} and corresponding \emph{attributes} used to perform the proposed quantitative analysis in this work. Also, Table~\ref{tab:concept} includes the class-wise expected criteria for each \emph{concept}.

The \emph{attributes} of the nuclei in a $\troi$ are computed as presented in Table~\ref{tab:concept}. It uses the $\troi$ and corresponding nuclei segmentation map, denoted as $I_{\text{seg}}$.
% define the area: A(x)
Area of a nucleus $x$, denoted as $A(x)$, is defined as the number of pixels belonging to $x$ in $I_{\text{seg}}$.
% define the perimeter and convex hull perimeter: P(x), P_CH(x)
$P(x)$, the perimeter of $x$, is measured as the contour length of $x$ in $I_{\text{seg}}$.
$P_{\text{ConvHull}}(x)$, the convex hull perimeter of $x$, is defined as the contour length of convex hull induced by $x$ in $I_{\text{seg}}$.
% define the minor and major axis length
The major and minor axis of $x$, noted as $a_{\text{major}}(x)$ and $a_{\text{minor}}(x)$, are the longest diameter of $x$ and the longest line segment perpendicular to $a_{\text{major}}(x)$, respectively. 
% define the GLCM matrix
The chromatin \emph{attributes} are computed from the normalized gray level co-occurrence matrix (GLCM)~\cite{Haralick1973}, which captures the probability distribution of co-occurring gray values in $x$.

\section*{Quantitative assessment}
In this section, we analyze two key components of the proposed quantitative metrics: the histogram construction and class separability scores for threshold set $\kset$. Furthermore, we relate the analysis to the class-wise expected criteria for each \emph{concept} presented in Table~\ref{tab:concept}.

\subsection*{Histogram analysis}
\begin{figure*}[!t]
\centering
\centerline{\includegraphics[width=0.91\linewidth]{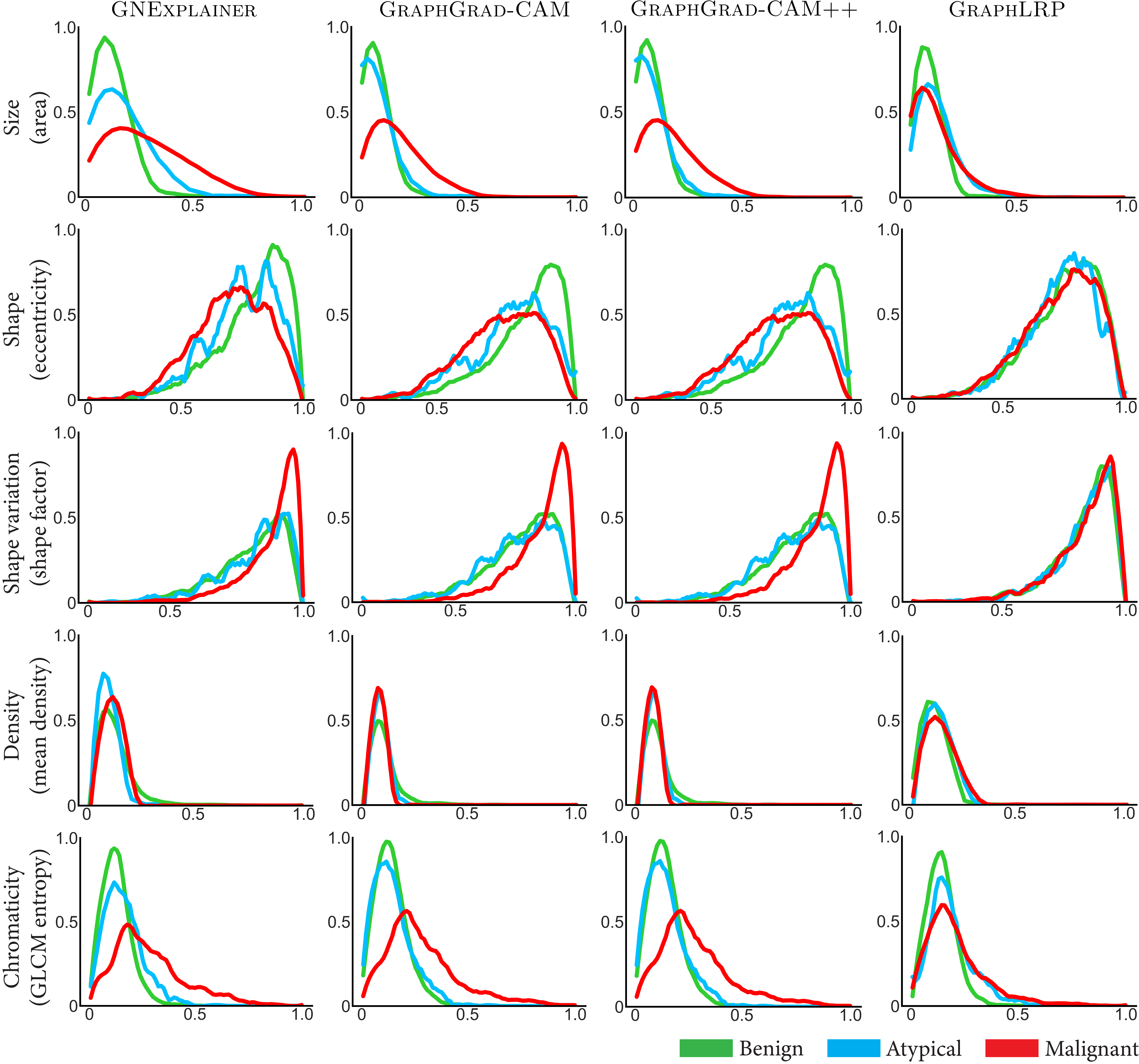}}
\caption{Per-class histograms for different \emph{concepts} across different graph explainers. For simplicity, histograms are presented for the best \emph{attribute} per \emph{concept} at fixed importance threshold $k=25$.}
\label{fig:histograms}
\end{figure*}

Histogram construction is a key component in the proposed quantitative metrics. Figure~\ref{fig:histograms} presents per-class histograms for each explainer and the best \emph{attribute} per \emph{concept}. We set the importance threshold to $k=25$, \ie for each $\troi$, we select 25 nuclei with the highest node importance. The best \emph{attribute} for a \emph{concept} is the one with the highest average pair-wise class separability.

The row-wise observation exhibits that $\gnnexplainer$ and $\graphlrp$ provide, respectively, the maximum and the minimum pair-wise class separability. The histograms for a \emph{concept} and for an explainer can be analyzed to assess the agreement between the selected important nuclei \emph{concept}, and the expected \emph{concept} behavior as presented in Table~\ref{tab:concept}, for all the classes.
For instance, nuclear \emph{area} is expected to be higher for malignant $\troi$s than benign ones. The \emph{area} histograms for $\gnnexplainer$, $\graphgradcam$ and $\graphgradcampp$ indicate that the important nuclei set in malignant $\troi$s includes nuclei with higher area compared to benign $\troi$s. 
Similarly, the important nuclei in malignant $\troi$s are expected to be vesicular, \ie high texture entropy, compared to light euchromatic, \ie moderate texture entropy, in benign $\troi$s. The \emph{chromaticity} histograms for $\gnnexplainer$, $\graphgradcam$ and $\graphgradcampp$ display this behavior.
%The considerable overlap between the histograms is due to the tumor heterogeneity.
Additionally, the histogram analysis can reveal the important \emph{concepts} and important \emph{attributes}. For instance, nuclear \emph{density} proves to be the least important \emph{concept} for differentiating the classes.

\subsection*{Separability score for threshold set $\kset$}
\begin{figure*}[!t]
\centering
\centerline{\includegraphics[width=0.87\linewidth]{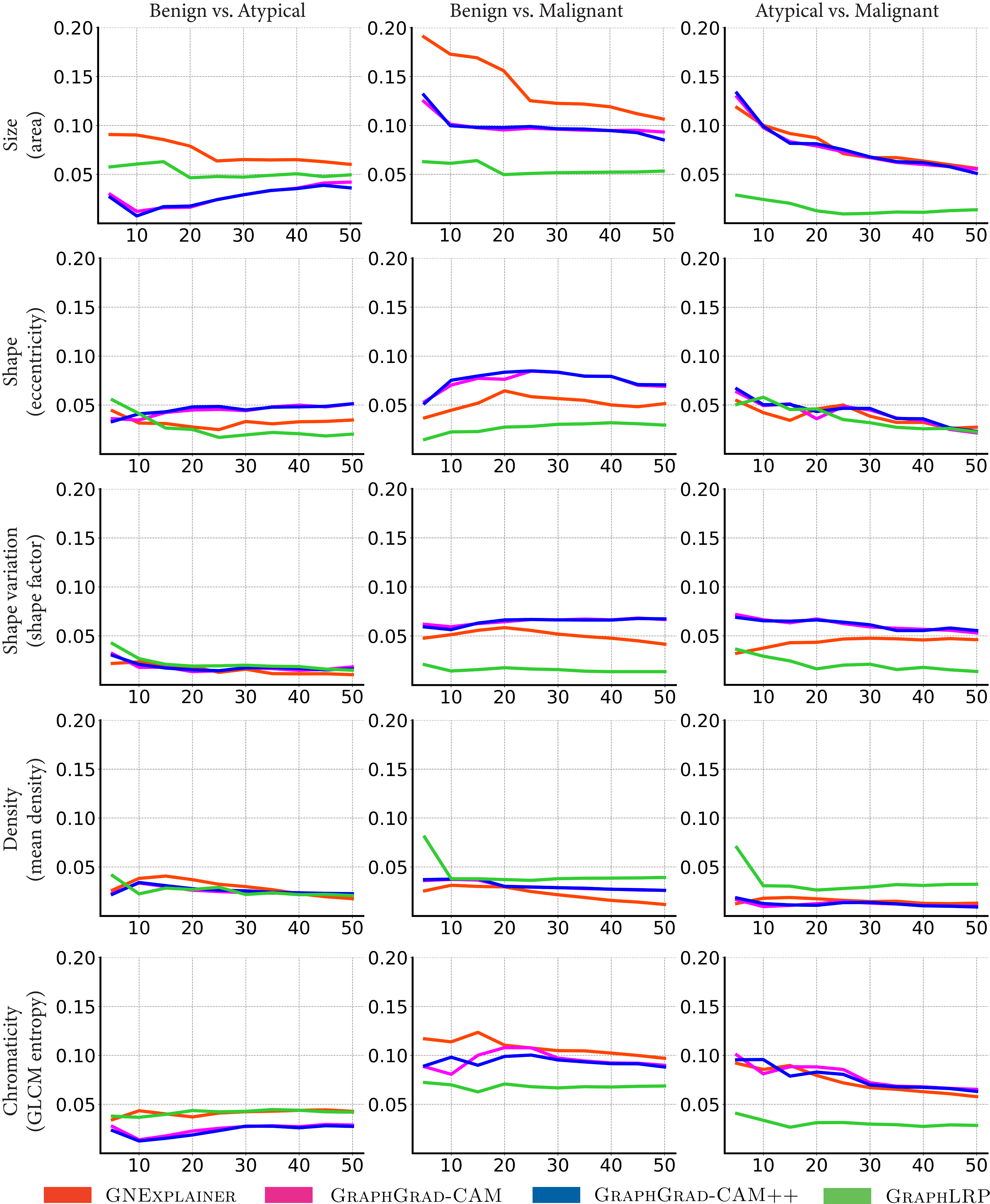}}
\caption{Visualizing the variation of pair-wise class separability score (Y-axis) w.r.t.\ various nuclei importance thresholds in $\kset$ (X-axis). The analysis is provided for different graph explainers, and for the best \emph{attribute} per \emph{concept}.}
\label{fig:kset}
\end{figure*}

Multiple importance thresholds $\kset$ are required to address the varying notion of important nuclei across different cell graphs and different explainers. Figure~\ref{fig:kset} presents the behavior of pair-wise class separability for using various $k \in \kset = \{5, 10, ..., 50 \}$. For simplicity, we present the behavior for the best \emph{attribute} per \emph{concept}.
In general, the pair-wise class separability is observed to decrease with decreasing $k$. Intuitively, decreasing $k$ results in including more unimportant nuclei into the evaluation, thereby gradually decreasing the class separability. 

The degree of agreement between the difference in the expected behavior per \emph{concept} and the pair-wise class separability in Figure~\ref{fig:kset}, for all pair-wise classifications and various $k \in \kset$ can be used to assess the explainer's quality.
For instance, according to Table \ref{tab:concept}, the difference in the expected nuclear \emph{size} can be considered as benign--atypical $<$ benign--malignant, and atypical--malignant $<$ benign--malignant. $\gnnexplainer$, $\graphgradcam$ and $\graphgradcampp$ display these behaviors $\forall k \in \kset$. $\gnnexplainer$ provides the highest class separability in each pair-wise classification, thus proving to be the best explainer pertaining to \emph{size concept}.
Detailed inspection of Figure~\ref{fig:kset} shows that all the differences in the expected behavior, per \emph{concept} for all pair-wise classifications, is inline with the \emph{concept}-wise expected behavior in Table~\ref{tab:concept}, $\forall c \in \conceptset$ and $\forall k \in \kset$.
Overall, $\gnnexplainer$ is seen to be the best explainer as it agrees to the majority of the expected differences $\forall c \in \conceptset$ for all pair-wise classifications, while providing high-class separability.  
Furthermore, \emph{size} proves to be the most important \emph{concept} that provides the maximum class separability across all pair-wise classifications.

\section*{Qualitative assessment}
\begin{figure*}[!t]
\centering
\centerline{\includegraphics[width=\linewidth]{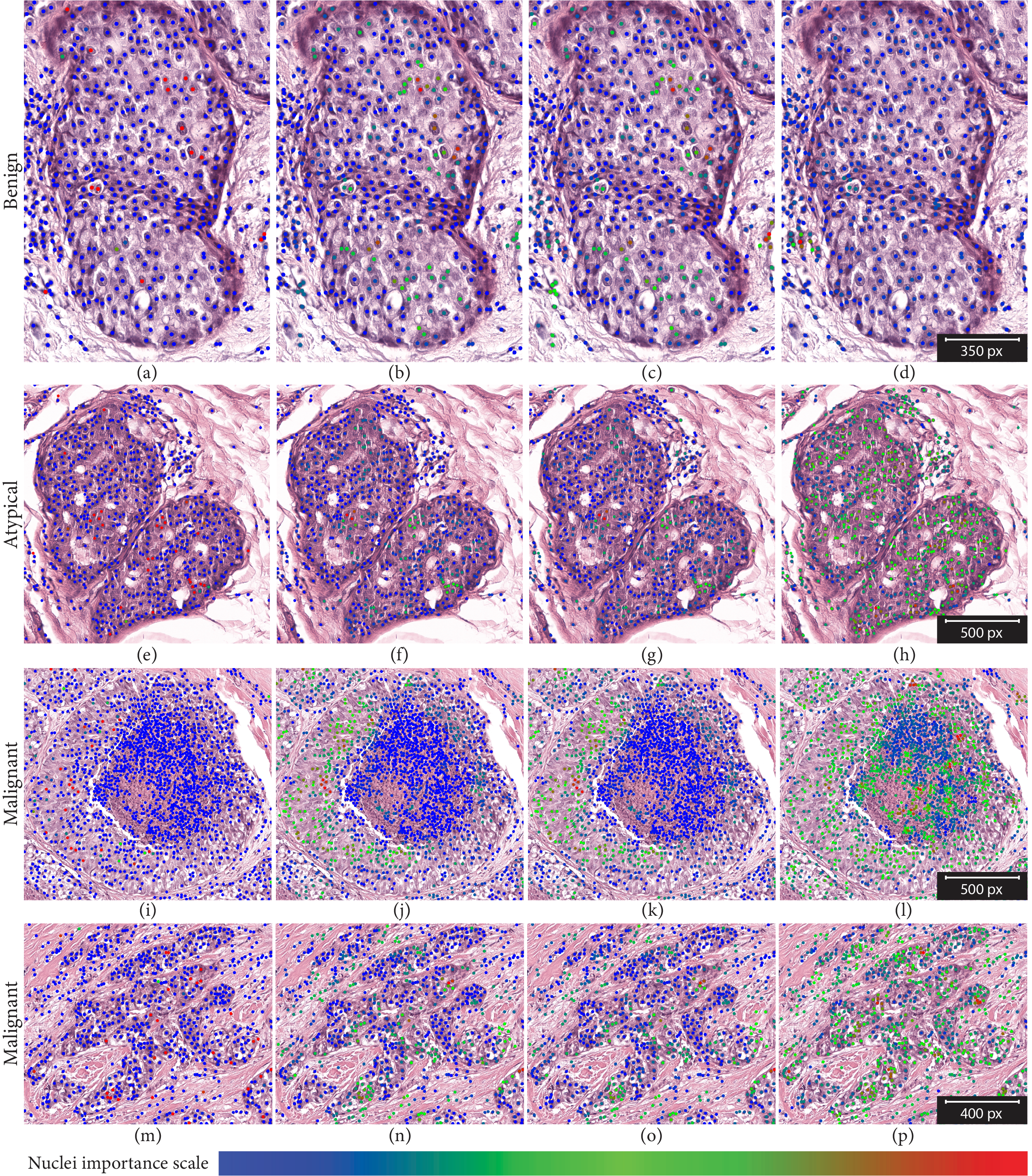}}
\caption{Qualitative results. The rows represent breast cancer subtypes, and columns represent graph explainers, \ie $\gnnexplainer$, $\graphgradcam$, $\graphgradcampp$, and $\graphlrp$. Nuclei level importance ranges from blue (the least important) to red (the highest important).}
\label{fig:explanations1}
\end{figure*}

\begin{figure*}[!t]
\centering
\centerline{\includegraphics[width=\linewidth]{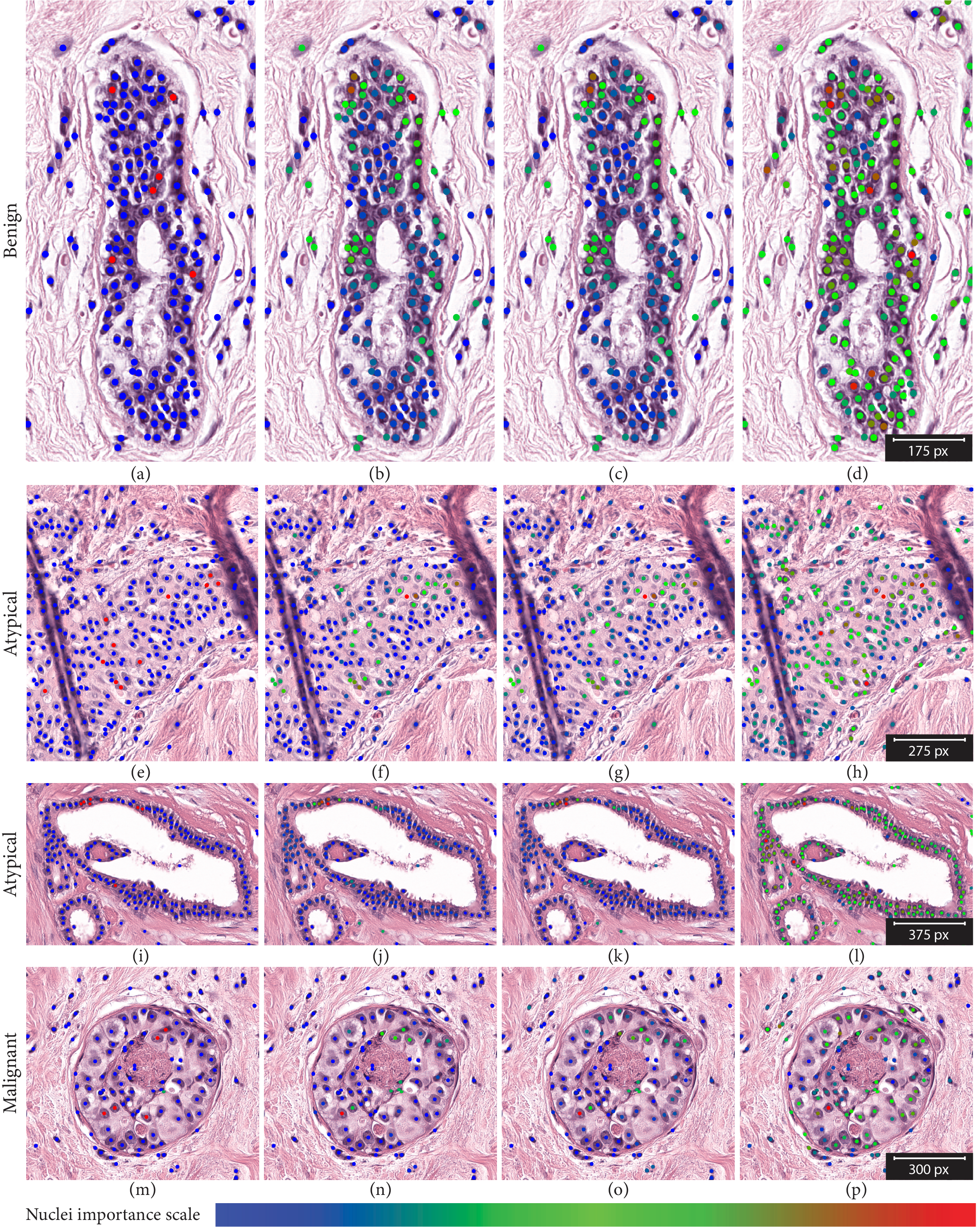}}
\caption{Qualitative results. The rows represent breast cancer subtypes, and columns represent graph explainers, \ie $\gnnexplainer$, $\graphgradcam$, $\graphgradcampp$, and $\graphlrp$. Nuclei level importance ranges from blue (the least important) to red (the highest important).}
\label{fig:explanations2}
\end{figure*}

Figure~\ref{fig:explanations1} and Figure~\ref{fig:explanations2} present $\cg$ explanations produced by $\gnnexplainer$, $\graphgradcam$, $\graphgradcampp$ and $\graphlrp$ for $\troi$s across benign, atypical and malignant breast tumors. It can be observed that $\gnnexplainer$ learns to binarize the explanations, thereby producing the most compact explanations by retaining the most important nuclei set of nuclei with high importance. However, $\graphgradcam$ and $\graphgradcampp$ produce explanations with more distributed nuclei importance than $\gnnexplainer$. $\graphlrp$ produces the largest explanations by retaining most of the nuclei in the $\cg$s.

\end{document}